\documentclass[journal]{IEEEtran}

\IEEEoverridecommandlockouts                              % This command is only needed if 
\title{
Adaptive Evolution‐Factor Risk–Ellipse Framework for Reliable and Safe Autonomous Driving
}

\author{Fujiang Yuan$^{1 \dagger}$, 
        Zhen Tian$^{2 \dagger}$, 
        Yangfan He$^{3}$, 
        Guojian Zou$^{4}$, 
        Chunhong Yuan$^{5}$, 
        Yanhong Peng$^{1,*}$, and Zhihao Lin$^{2,*}$%

\thanks{$^{1}$College of Mechanical Engineering, Chongqing University of Technology, Chongqing, 400054, China.}%
\thanks{$^{2}$James Watt School of Engineering, University of Glasgow, Glasgow G12 8QQ, United Kingdom.}%
\thanks{$^{3}$ Computer science with the University of Minnesota - Twin Cities, Minneapolis, MN, USA.}%
\thanks{$^{4}$ College of Transportation Engineering, Tongji University, Shanghai, 201804,  China.}
\thanks{$^{5}$Laboratory of Intelligent Home Appliances, College of Science and Technology, Ningbo University, Ningbo 315300, Zhejiang, China.}%
\thanks{*Corresponding author.}%
\thanks{$\dagger$ Equal contribution}
}
\usepackage{xcolor}

\usepackage{hyperref}
\usepackage{cite}
\usepackage{algorithm}  
\usepackage{algpseudocode}
\usepackage{multirow}
\usepackage{amsmath}
\usepackage{amssymb}
\usepackage{epsfig} % for postscript graphics files

\usepackage{amsthm}

\usepackage{graphicx}
\usepackage{caption}
\usepackage{subcaption}%小表
\usepackage{enumitem} %自定义item

\usepackage[flushleft]{threeparttable}%加表格描述用的

\usepackage{tabularx} % 使用tabularx包来灵活控制表格宽度

\usepackage{titlesec}

\begin{document}

\maketitle

%%%%%%%%%%%%%%%%%%%%%%%%%%%%%%%%%%%%%%%%%%%%%%%%%%%%%%%%%%%%%%%%%%%%%%%%%%%%%%%%
\begin{abstract}
In recent years, ensuring safety, efficiency, and comfort in interactive autonomous driving has become a critical challenge. Traditional model-based techniques, such as game-theoretic methods and robust control, are often overly conservative or computationally intensive. Conversely, learning-based approaches typically require extensive training data and frequently exhibit limited interpretability and generalizability. Simpler strategies, such as Risk Potential Fields (RPF), provide lightweight alternatives with minimal data demands but are inherently static and struggle to adapt effectively to dynamic traffic conditions. To overcome these limitations, we propose the Evolutionary Risk Potential Field (ERPF), a novel approach that dynamically updates risk assessments in dynamical scenarios based on historical obstacle proximity data. We introduce a Risk-Ellipse construct that combines longitudinal reach and lateral uncertainty into a unified spatial–temporal collision envelope. Additionally, we define an adaptive Evolution Factor metric, computed through sigmoid normalization of Time-to-Collision (TTC) and Time-Window-of-Hazard (TWH), which dynamically adjusts the dimensions of the ellipse axes in real time. This adaptive risk metric is integrated seamlessly into a Model Predictive Control (MPC) framework, enabling autonomous vehicles to proactively address complex interactive driving scenarios in terms of uncertain driving of surrounding vehicles. Comprehensive comparative experiments demonstrate that our ERPF-MPC approach consistently achieves smoother trajectories, higher average speeds, and collision-free navigation, offering a robust and adaptive solution suitable for complex interactive driving environments.
\end{abstract}


%%%%%%%%%%%%%%%%%%%%%%%%%%%%%%%%%%%%%%%%%%%%%%%%%%%%%%%%%%%%%%%%%%%%%%%%%%%%%%%%
\section{INTRODUCTION}

%, and their numbers are predicted to exceed 50 million by 2024~\cite{ignatious2022overview}.
\IEEEPARstart{I}n recent years, autonomous driving has advanced rapidly. Many research efforts have focused on self-driving technology. The autonomous driving includes perception~\cite{lin2024enhanced,lin2024dpl,lin2025slam2}, planning~\cite{reda2024path,yuan2025bio,zuo2025industrial,zuo2025advanced,chen2024end,teng2023motion,mao2024octomap,li2025efficient}, and control~\cite{tsai2024autonomous,barruffo2024goa4,liu2025ensuring,chen2025control}. However, interactive driving scenarios still pose significant challenges in planning. Safety remains the top priority. In Figure 1, an autonomous vehicle (AV) shares the road with several human-driven vehicles (HDVs), creating a zone where a collision could occur if proper caution is not taken. The AV must continuously assess potential risks and plan its trajectory to avoid collisions while navigating among other vehicles.

\begin{figure}[t]
      \centering
      \includegraphics[width=\linewidth]{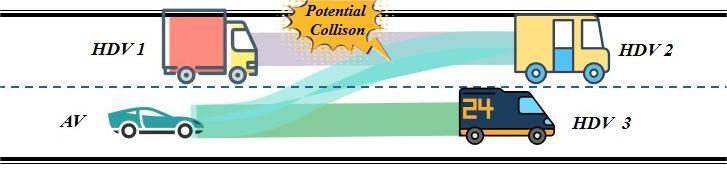}
      \caption{Illustration of safe driving in modern and intelligent city}
      \label{figure2}
    \end{figure}

Model-based methods have been widely explored. For example, game-theoretic approaches attempt to predict other drivers' behaviors~\cite{qin2024game,zhang2024non,zheng2025enhanced,zheng2025mean,liu2025aco}. They assume a specific driving style for opponents. This assumption can be different from real situation. Therefore, these methods often lead to overly conservative decisions. Robust control techniques~\cite{wang2024robust} are another model-based approach. They enlarge the feasible set to account for uncertainty. Unfortunately, this expansion may include regions that are actually dangerous.

Learning-based methods have also attracted much attention~\cite{lin2024conflicts,10607945,tian2025evaluating,zhen2025balanced}. They rely on large amounts of training data. Therefore, these methods require substantial computational resources. Moreover, learning-based approaches often operate as black boxes~\cite{hassija2024interpreting}, leading their decision-making process lacks transparency.

Risk Potential Field (RPF) methods~\cite{tan2021risk}, in contrast, offer lightweight and interpretable alternatives by modeling repulsive fields around obstacles. However, conventional RPFs are static, fail to adapt to evolving scenarios, and lack explicit temporal modeling, which restricts their effectiveness in high-speed or congested situations.

To bridge these gaps, we propose a novel Evolutionary Risk Potential Field (ERPF) framework, which integrates historical proximity data and risk evolution into a dynamic, interpretable potential field. ERPF adapts risk levels based on recent vehicle interactions and is further fused into a Model Predictive Control (MPC) formulation, forming the proposed ERPF-MPC system. Unlike conventional methods, our framework proactively reshapes risk fields in real-time and leverages time-aware risk ellipses governed by Time-to-Collision (TTC) and Time-Window-of-Hazard (TWH). These ellipses are continuously adapted using a sigmoid-normalized evolution factor, enabling early warning, smooth avoidance, and computational feasibility.
\begin{figure*}[t]
      \centering
      \includegraphics[width=0.75\linewidth]{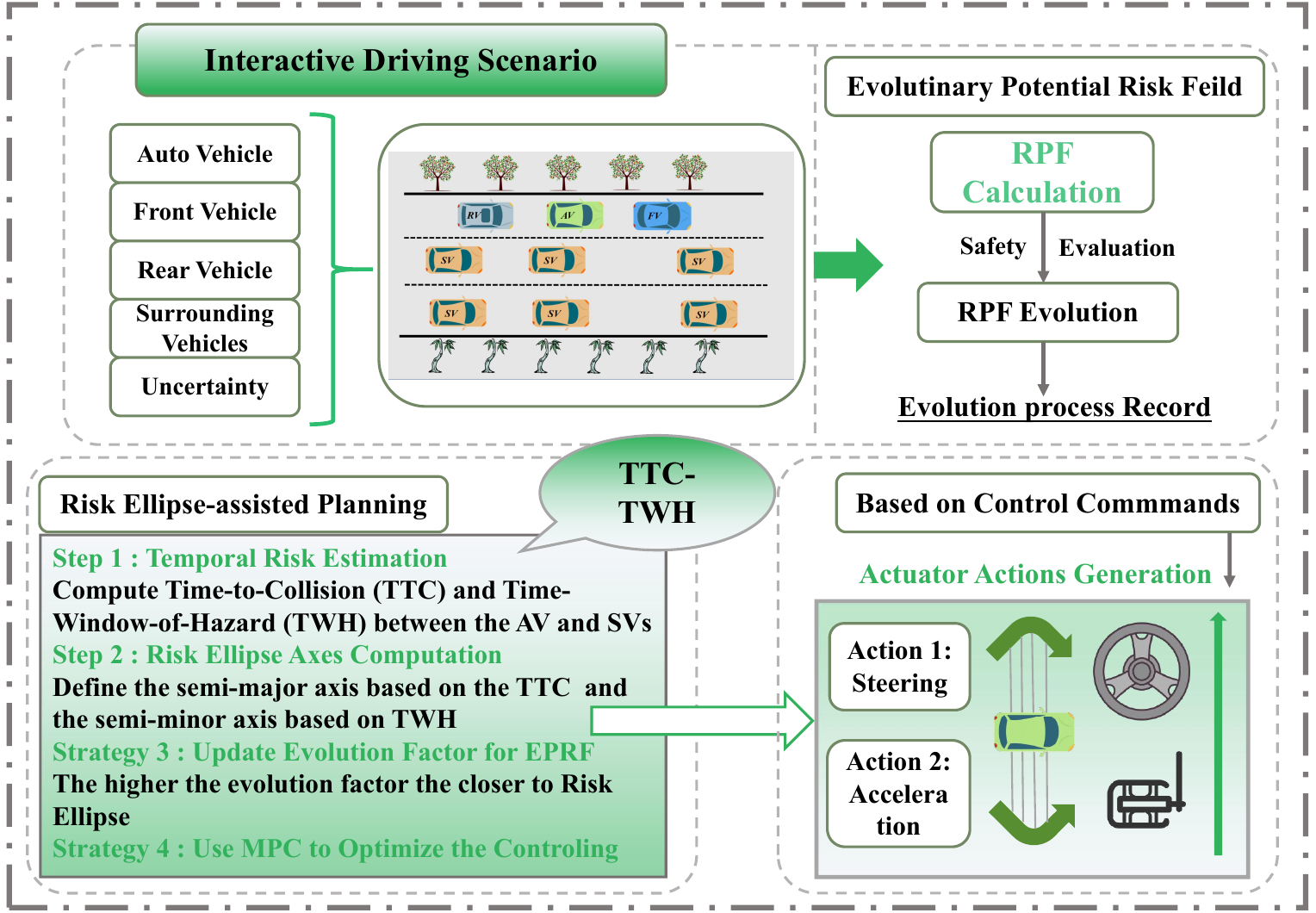}
      \caption{The proposed System Workflow using in smart cities.}
      \label{figure2}
\end{figure*}
The main contributions of this work are: \begin{itemize}
    \item We propose a novel EPRF that dynamically adapts to interactive driving scenarios using historical proximity measurements. This design addresses the inflexibility and high computational cost of traditional Risk Potential Fields while preserving interpretability.
    
    \item We integrate ERPF with MPC to construct a unified ERPF-MPC planning framework. This framework enables real-time risk-aware decision-making via a newly defined risk ellipse and adaptive evolution factor, which jointly model spatial–temporal collision risk based on Time-to-Collision (TTC) and Time-Window-of-Hazard (TWH).
    
    \item We conduct extensive simulation-based evaluation in typical interactive driving scenarios such as lane changing and overtaking. Results demonstrate that our method improves safety (collision avoidance), driving efficiency (higher average speed), and ride comfort (smoother trajectories) over state-of-the-art baselines.
\end{itemize}

In the following sections, we introduce the related works, the system model, the risk field formulation, the evolution mechanism, and the MPC optimization problem. Experimental results validate the effectiveness of our proposed approach in interactive driving scenarios.

\section{RELATED WORKS}

\subsection{Challenges in Interactive Autonomous Driving}
Interactive autonomous driving poses many challenges. In dynamic traffic environments, vehicles must operate safely while interacting with human driven vehicles (HDVs)~\cite{liu2023designing,liang2024interaction,gong2022modeling,lin2025safetyr,lin2025safety,lin2025multi}. The risks of collisions raises during the interactive driving, due to the late steering or judgment errors. Moreover, interactive scenarios such as lane changes and overtaking require fast and reliable decision making~\cite{yao2023optimal,chai2022multiphase,palatti2021planning}. The vehicle must balance safety, efficiency, and ride comfort. Many systems struggle to achieve these goals simultaneously. As a result, ensuring safety in complex and uncertain traffic conditions remains a critical challenge.

\subsection{Shortcomings of Current Planning Methods}
Several planning methods have been proposed to address autonomous driving tasks. However, each approach has its limitations.

MPC is a popular optimization-based method~\cite{lazcano2021mpc,zhai2022mpc,11041353}. It predicts future vehicle states by solving a finite-horizon optimal control problem. MPC is powerful and flexible. Yet, it heavily relies on accurate models and predictions of the environment~\cite{tian2025risk}. In interactive scenarios, uncertainty in obstacle behavior can degrade its performance. Moreover, some adapted MPC is computationally demanding~\cite{fu2022nmpc}. 

Control barrier functions (CBFs) are designed to enforce safety constraints~\cite{ames2019control,liu2023auxiliary,liu2023iterative,liu2024learning,liu2025auxiliary,liu2025risk,liu2025ensuring}.  They define a barrier function that prevents the system from entering unsafe states. A CBF for avoiding dynamic objectives is proposed in~\cite{9483029}. While CBFs provide formal safety guarantees, they can be overly conservative. The design of the barrier function may force the vehicle to yield excessively. This can reduce overall driving efficiency. In addition, the trade-off between safety and performance is not always easy to balance using CBF methods.

Risk potential fields compute a risk value based on the distance between the vehicle and obstacles~\cite{triharminto2016novel,li2025adaptive}. They offer a simple way to guide the vehicle away from danger. In \cite{yao2020path}, a framework integrating the RPF method with reinforcement learning is introduced to enable collision avoidance in environments with dense obstacles. However, traditional RPF methods use fixed parameters. They do not adapt to changes in the environment. The computed risk may not reflect the true level of danger. Moreover, some formulations require complex computations. This can increase the computational burden and limit real-time performance.

In summary, while existing model-based and learning-based approaches have advanced autonomous driving, they suffer from key drawbacks. Model-based methods like MPC and CBF are either computationally intensive or too conservative. Learning-based methods need extensive training and often lack transparency. In contrast, simpler guidance methods such as RPF are computationally efficient but lack adaptability. Our work addresses these issues by proposing a self-evolving risk potential field that can adapt to the dynamic environment. When integrated with MPC, ERPF-MPC achieves improved safety, efficiency, and comfort in interactive driving scenarios.

\section{SYSTEM OVERVIEW}

Figure~\ref{figure2} 
illustrates the proposed ERPF system workflow for intelligent autonomous driving. The framework processes interactive driving scenarios involving multiple vehicles (ego, front, rear, and surrounding vehicles) through a three-stage pipeline. First, the system computes basic RPF values, performs safety evaluation, and evolves the risk field based on historical proximity data. Second, the risk ellipse-assisted planning module calculates TTC and TWH parameters, defines ellipse axes accordingly, updates evolution factors for ERPF, and optimizes control through MPC. Finally, the system generates actuator commands for steering and acceleration. This integrated approach enables real-time risk-aware decision-making by combining spatial-temporal collision modeling with predictive control, ensuring safe navigation in complex multi-agent environments.

%%%%%%%%%%%%%%%%%%%%%%%%%%%%%%%%%%%%%%%%%%%%%%%%%%%%%%%%%%%%%%%
\section{Evolutionary Potential Risk Field-based Planning}
\subsection{System Dynamics and Driving Scenario}
We consider the host vehicle as a point-mass moving in a planar environment. Its state vector is defined as
\begin{equation}
s_k = \begin{bmatrix} x_k \\ y_k \\ v_k \end{bmatrix} \in \mathbb{R}^{3},
\end{equation}
where \(x_k\) and \(y_k\) denote the position coordinates and \(v_k\) is the longitudinal velocity. A discrete-time kinematic model is given by
\begin{equation} \label{eq:dynamics}
s_{k+1} = A s_k + B u_k,
\end{equation}
with the control input
\begin{equation}
u_k = \begin{bmatrix} a_k \\ v_{y,k} \end{bmatrix} \in \mathbb{R}^{2},
\end{equation}
where \(a_k\) is the longitudinal acceleration and \(v_{y,k}\) is the lateral velocity command. A simple linearized model over a sampling period \(\Delta t\) is
\begin{equation}
A = \begin{bmatrix} 1 & 0 & \Delta t \\ 0 & 1 & 0 \\ 0 & 0 & 1 \end{bmatrix}, \quad
B = \begin{bmatrix} 0 & 0 \\ 0 & \Delta t \\ \Delta t & 0 \end{bmatrix}.
\end{equation}

For driving scenarios, a lane-change maneuver is considered where the desired lateral (y-axis) reference changes from one lane center \(y_1\) to another \(y_2\). The reference trajectory is defined as
\begin{equation}
s_k^{\mathrm{ref}} = \begin{bmatrix} x_k^{\mathrm{ref}} \\ y_k^{\mathrm{ref}} \\ v_{\mathrm{ref}} \end{bmatrix}, \quad y_k^{\mathrm{ref}} = y_1 + \frac{k}{N}(y_2 - y_1),
\end{equation}
and an overtaking scenario is similarly described. In addition, multiple obstacle vehicles are present. For each obstacle \(i\) (\(i=1,\ldots,N_{\text{obs}}\)), its (predicted) position is denoted by
\[
p_i(k) \in \mathbb{R}^2, \quad \text{with } p_i(k) = p_i(0) + \mathbf v_i\, k \Delta t,
\]
where \(\mathbf v_i = \begin{bmatrix} v_{x,i} \\ v_{y,i} \end{bmatrix}\) is the 2D velocity of obstacle \(i\).
\begin{figure}[t]
      \centering
      \includegraphics[width=1\linewidth]{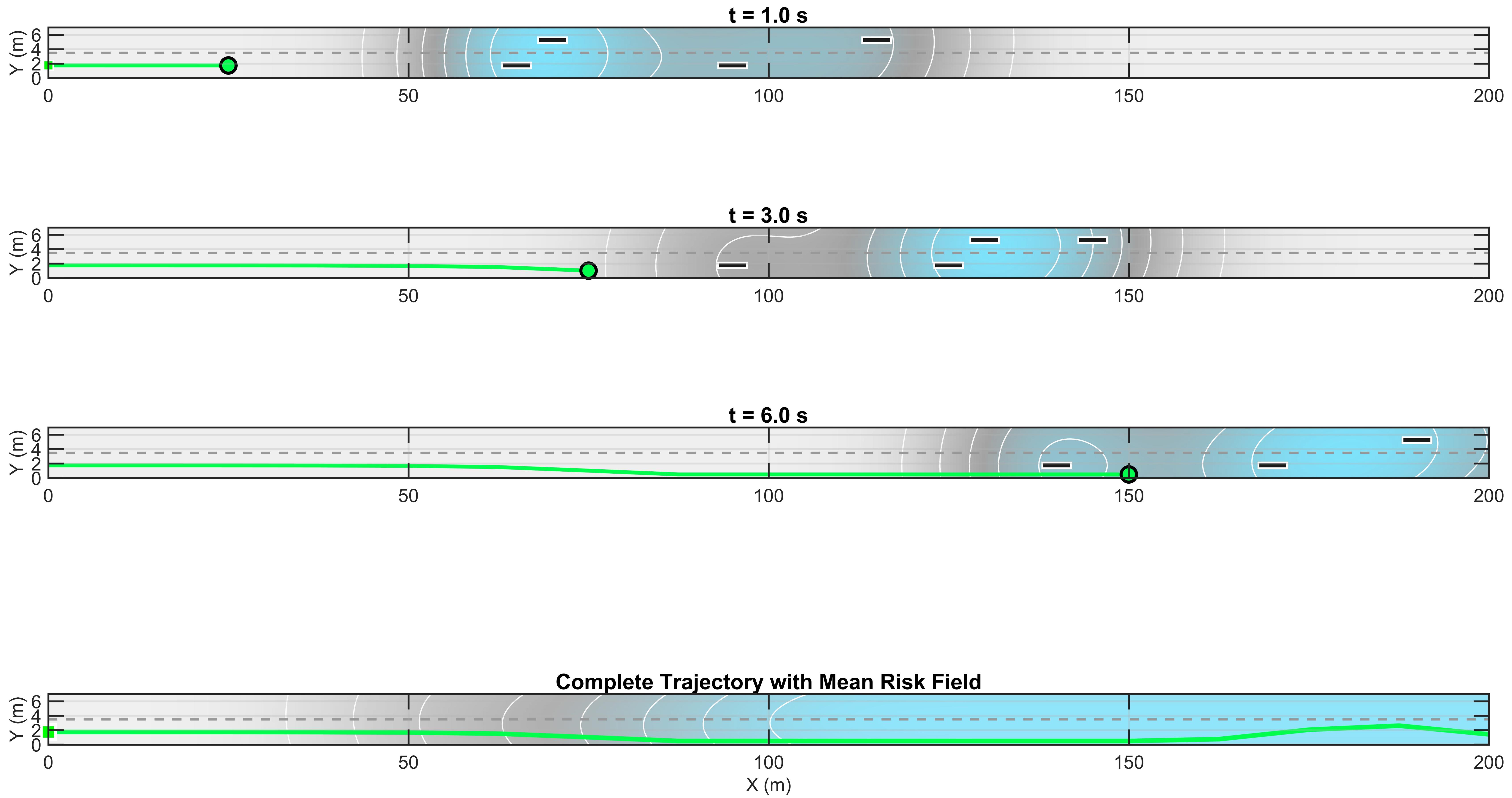} %inter
      \caption{Evolution of the ERPF and corresponding vehicle trajectory in a lane‐change scenario, with risk intensity shown from low to high.}
      \label{inner}
\end{figure}

\subsection{Evolutionary Risk Potential Field}
We define the standard RPF for obstacle \(i\) as a function of the Euclidean distance between the host vehicle and the obstacle. Let
\[
d_i(s_k) = \| C s_k - p_i(k) \|,
\]
where \(C = \begin{bmatrix} I_2 & \mathbf{0}_{2 \times 1} \end{bmatrix}\) extracts the position from \(s_k\). The risk function is then given by
\begin{equation} \label{eq:rpf}
\phi(d) = \begin{cases}
\displaystyle \frac{1}{\max(d, \epsilon)} - \frac{1}{d_{\text{safe}}}, & d < d_{\text{safe}}, \\[1ex]
0, & d \ge d_{\text{safe}},
\end{cases}
\end{equation}
where $\epsilon = 0.1$ m is a small constant to ensure numerical stability, and the RPF for obstacle \(i\) is
\begin{equation}
V_\text{RPF}^i(s_k) = \alpha_i \, \phi\bigl(d_i(s_k)\bigr),
\end{equation}
with \(\alpha_i\) as a gain. The total risk is aggregated as
\begin{equation}
V_\text{RPF}(s_k) = \sum_{i=1}^{N_{\text{obs}}} V_{RPF}^i(s_k).
\end{equation}

To capture the evolution of risk based on historical obstacle proximity, we define a history measure for each obstacle. Let the average distance over a history window of \(N_H\) steps be
\begin{equation}
\bar{d}_i(k) = \frac{1}{N_H} \sum_{j=k-N_H+1}^{k} d_i(s_j).
\end{equation}
The evolutionary factor is then defined by
\begin{equation} \label{eq:evolution_smooth}
\eta_i(k) = 1 + \lambda \cdot \text{sigmoid}\left( \frac{\bar{d}_i(k) - d_i(s_k)}{d_{\text{safe}}} \right)
\end{equation}
where \(\lambda > 0\) is a design parameter. When $\bar{d}_i(k) > d_i(s_k)$, the obstacle is approaching and the risk is amplified. The Evolutionary Risk Potential Field is given by
\begin{equation}
V_\text{ERPF}(s_k) = \sum_{i=1}^{N_{\text{obs}}} \eta_i(k) \, V_\text{RPF}^i(s_k).
\end{equation}

Define the state and control sequences over the prediction horizon \(N\) as
\[
S = \begin{bmatrix} s_0 \\ s_1 \\ \vdots \\ s_N \end{bmatrix} \in \mathbb{R}^{(N+1)n}, \quad U = \begin{bmatrix} u_0 \\ u_1 \\ \vdots \\ u_{N-1} \end{bmatrix} \in \mathbb{R}^{Nm}.
\]
By iterating the dynamics \eqref{eq:dynamics}, we have the stacked constraint
\begin{equation}
S = \mathcal{A} s_0 + \mathcal{B} U,
\end{equation}
where the matrices \(\mathcal{A}\) and \(\mathcal{B}\) are defined as
\[
\mathcal{A} = \begin{bmatrix} I \\ A \\ A^2 \\ \vdots \\ A^N \end{bmatrix}, \quad \mathcal{B} = \begin{bmatrix} 0 & \cdots & 0 \\
B & \cdots & 0 \\
A B & \cdots & 0 \\
\vdots & \ddots & \vdots \\
A^{N-1} B & \cdots & B \end{bmatrix}.
\]

The cost function over the prediction horizon is formulated as
\begin{equation} \label{eq:cost}
\begin{aligned}
J(U) &= \sum_{k=0}^{N-1} \left\{ \|s_k - s_k^{\mathrm{ref}}\|_Q^2 + \|u_k\|_R^2 + \gamma\, V_\text{ERPF}(s_k) \right\} \\
&\quad + \|s_N - s_N^{\mathrm{ref}}\|_{Q_N}^2
\end{aligned}
\end{equation}
which can be compactly written in matrix form as
\begin{equation}
J(U) = \| \mathcal{B}U + \mathcal{A} s_0 - S^{\mathrm{ref}} \|_{\mathcal{Q}}^2 + \|U\|_{\mathcal{R}}^2 + \gamma \sum_{k=0}^{N-1} V_\text{ERPF}(s_k),
\end{equation}
\begin{equation}
\mathcal{Q} = \operatorname{diag}(Q, Q, \dots, Q, Q_N) \in \mathbb{R}^{(N+1)n \times (N+1)n},
\end{equation}
\begin{equation}
\mathcal{R} = \operatorname{diag}(R, R, \dots, R) \in \mathbb{R}^{Nm \times Nm}.
\end{equation}
where \(S^{\mathrm{ref}} = [s_0^{\mathrm{ref}\,T}, \ldots, s_N^{\mathrm{ref}\,T}]^T\). Here, \(\gamma\) is a positive weighting factor.
\begin{figure}[t]
      \centering
      \includegraphics[width=0.8\linewidth]{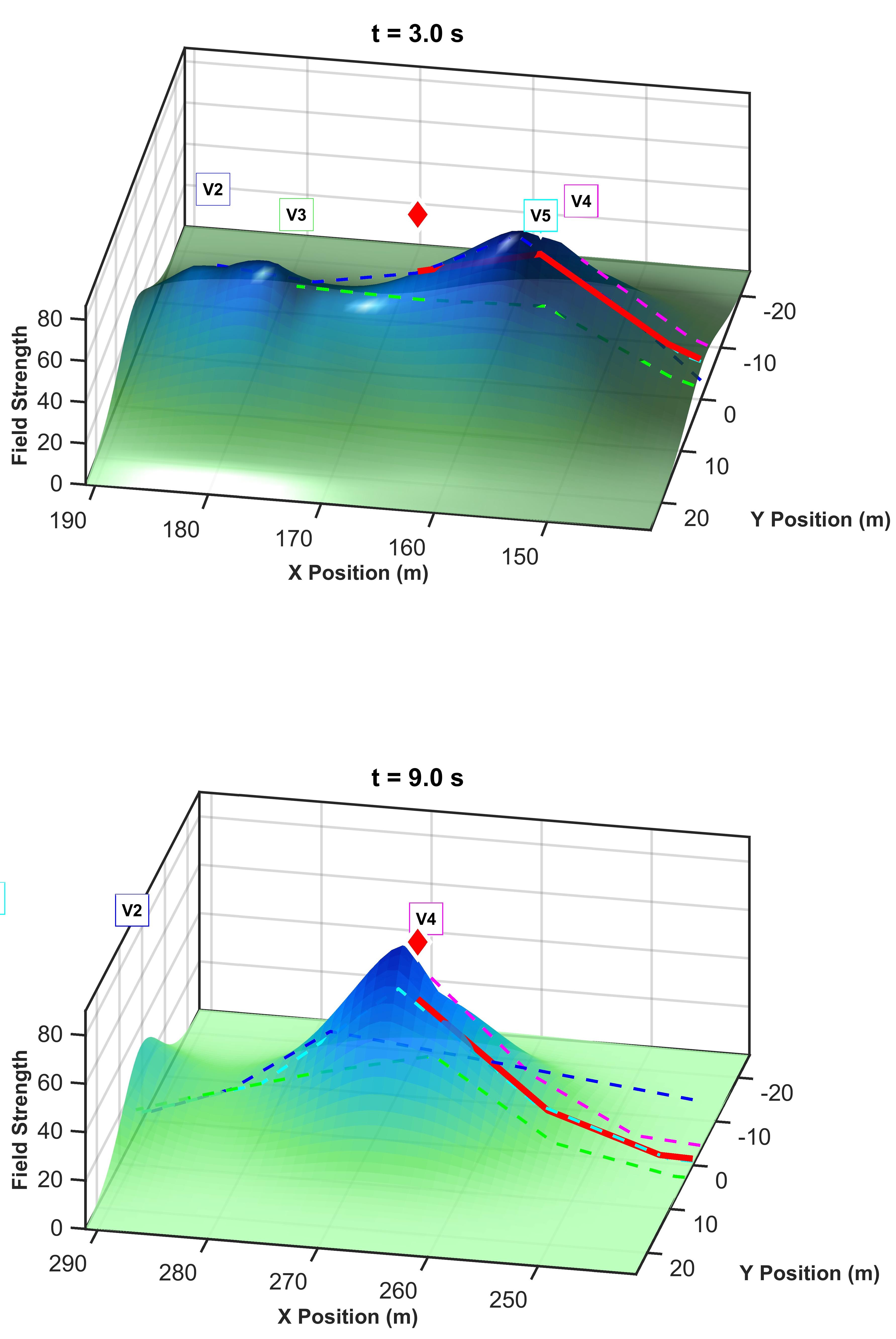}
      \caption{Time-varying ERPF field strength visualization at $t=3.0s$ and $t=9.0s$, illustrating dynamic adaptation to multi-agent interaction intensity.}
      \label{lt}
\end{figure}
Thus, the overall MPC optimization problem becomes
\begin{equation} \label{eq:mpc_optimization}
\begin{aligned}
\min_{U} \quad & J(U) \\
\text{s.t.} \quad & S = \mathcal{A} s_0 + \mathcal{B} U, \\
& U \in \mathcal{U}, \quad S \in \mathcal{S}.
\end{aligned}
\end{equation}

In practice, the risk term \(V_{ERPF}(s_k)\) is nonconvex and may be linearized or approximated iteratively.

Figure~\ref{inner} illustrates how the ERPF dynamically adapts to moving obstacles. The panels show snapshots at $t = 1$, $3$, and $6$ seconds, respectively, producing a smoothly evolving heatmap that accurately highlights emerging collision zones. By sampling the local lateral gradient of the ERPF, the vehicle continuously adjusts its lane position to avoid high-risk areas, resulting in the blue trajectory shown at the bottom, which progressively deviates around danger hotspots. The averaged field further reveals the overall safety envelope and confirms that the history-informed evolution in ERPF provides early warnings, stable gradient cues, and seamless integration with downstream control modules such as MPC. This ensures both proactive collision avoidance and smooth path generation.

Figure~\ref{lt} presents two snapshots of the ERPF field at different time instances, demonstrating its dynamic response to the interaction structure. The red trajectory corresponds to the ego vehicle (AV), while V2–V5 represent nearby human-driven vehicles exerting repulsive influence. At $t=3.0\,s$, the AV navigates a densely interactive region with multiple close-proximity vehicles (V3, V4, V5). The field exhibits two distinct peaks—indicating dual repulsion zones—corresponding to high interaction intensity. This configuration demands careful motion planning under compound risk influences. By contrast, at $t=9.0\,s$, the AV encounters a sparser scenario with only V4 exerting significant influence. Accordingly, the field adjusts to a single prominent peak, and the overall potential landscape becomes smoother.

\begin{figure}[t]
      \centering
      \includegraphics[width=1\linewidth]{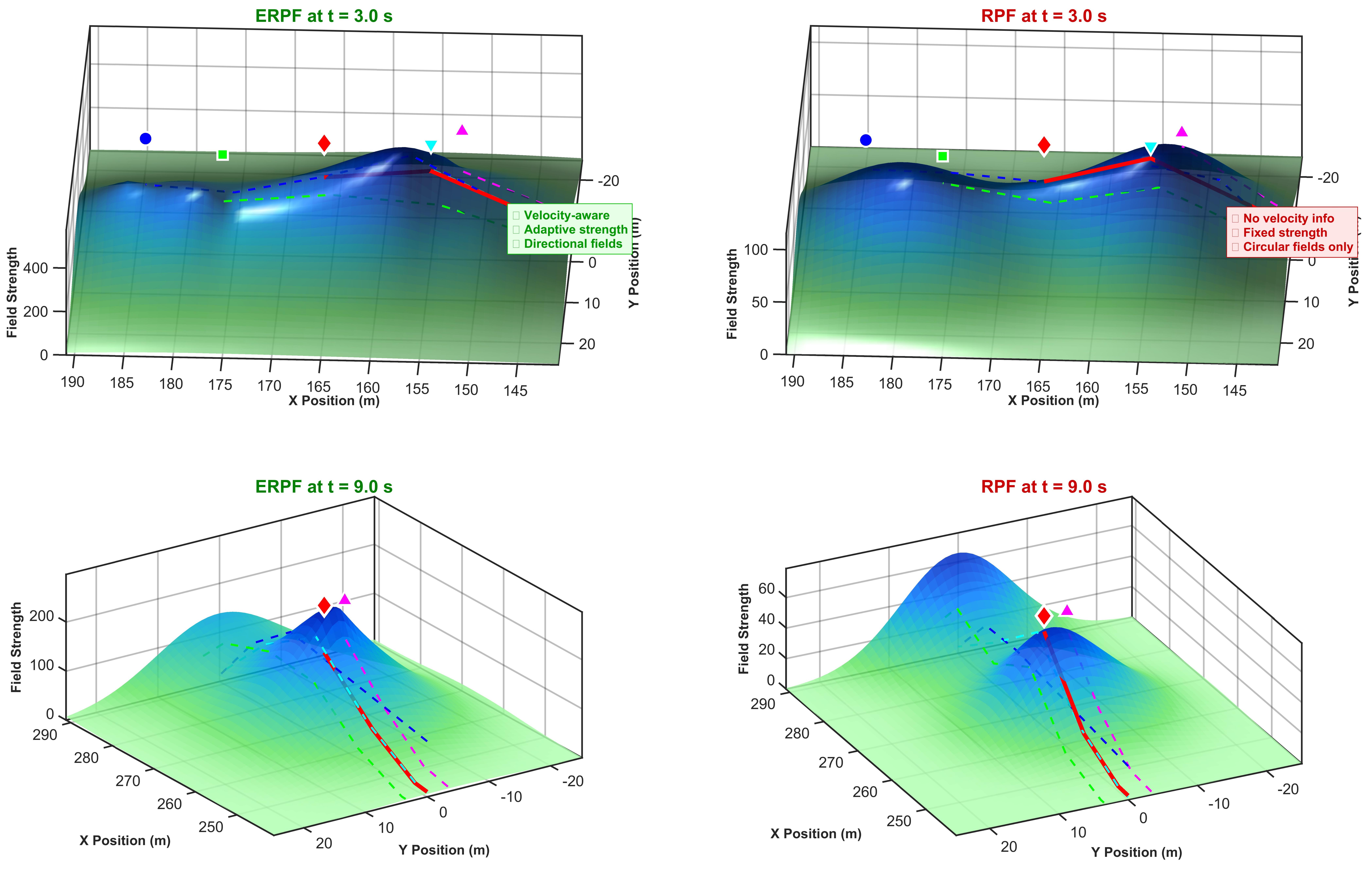}
      \caption{Comparison between ERPF and classical RPF at $t = 3.0s$ and $t = 9.0s$. ERPF captures multiple dynamic risk sources with directional and strength-adaptive modeling.}
      \label{ltd}
\end{figure}

Figure~\ref{ltd} compares the proposed Evolution-aware Risk Potential Field (ERPF) with the classical Risk Potential Field (RPF) framework at two critical moments: $t=3.0\,s$ and $t=9.0\,s$. At $t=3.0\,s$, multiple vehicles exert influence on the ego vehicle. The ERPF model (top-left) successfully captures two distinct risk peaks—one from the blue vehicle and another from the green vehicle—highlighting its superior ability to distinguish and localize concurrent interactive threats. In contrast, the RPF (top-right) merges their influence into a single blended peak, underestimating the spatial diversity and failing to reflect the true complexity of the interaction. At $t=9.0\,s$, as the leading vehicle has already moved far ahead, the ERPF (bottom-left) appropriately shifts its attention to nearby vehicles, generating a concentrated risk peak from the adjacent vehicle while ignoring distant entities. The RPF model (bottom-right), however, continues to display significant field strength toward the now-irrelevant forward vehicle, indicating a lack of dynamic filtering.

\subsection{Driving Scenarios with Obstacle Vehicles}
In a typical driving scenario such as a lane-change maneuver, the host vehicle is required to change from lane 1 to lane 2 while avoiding other moving vehicles. Let the lane centers be \(y_1\) and \(y_2\) respectively. The reference lateral trajectory is set as
\begin{equation}
y_k^{\mathrm{ref}} = y_1 + \frac{k}{N}(y_2 - y_1), \quad k=0,\dots,N.
\end{equation}
Obstacle vehicles are modeled as moving objects with positions
\begin{equation}
p_i(k) = p_i(0) + \mathbf v_i\, k \Delta t, \quad i = 1, \dots, N_{\text{obs}},
\end{equation}
where \(\mathbf v_i\) represents the 2D velocity of obstacle \(i\). The risk functions \eqref{eq:rpf} are computed for each obstacle, and the historical average distance \(\bar{d}_i(k)\) is updated in real time. In overtaking scenarios, additional constraints or modifications to the reference trajectory may be applied.

Let
\[
H = 2 \left( \mathcal{B}^\top \mathcal{Q} \mathcal{B} + \mathcal{R} \right), \quad g = 2\mathcal{B}^\top \mathcal{Q}\bigl(\mathcal{A}s_0 - S^{\mathrm{ref}}\bigr).
\]
Then, excluding the risk term, the tracking and control cost can be written in quadratic form as
\begin{equation}
J_{\text{track}}(U) = \frac{1}{2} U^\top H U + g^\top U.
\end{equation}
The complete optimization problem, including the risk term, is
\begin{equation}
\min_{U} \; \frac{1}{2} U^\top H U + g^\top U + \gamma \sum_{k=0}^{N-1} \sum_{i=1}^{N_{\text{obs}}} \eta_i(k) \, \phi\Bigl(\|Cs_k - p_i(k)\|\Bigr),
\end{equation}
subject to the linear dynamics constraint \(S = \mathcal{A} s_0 + \mathcal{B}U\) and state/input constraints \(S \in \mathcal{S}\), \(U \in \mathcal{U}\).

The overall ERPF-MPC procedure for autonomous driving is summarized in Algorithm~\ref{alg:ERPFMPC}.

\begin{algorithm}[t]
\caption{ERPF-MPC for Autonomous Driving}
\label{alg:ERPFMPC}
\begin{algorithmic}[1]
\State \textbf{Input:} Current state $s_0$, reference trajectory $S^{\mathrm{ref}}$, obstacle states $p_i(k)$ for $i=1,\dots,N_{\text{obs}}$, prediction horizon $N$, system matrices $A,B$, cost matrices $Q,R,Q_N$, risk parameters $\alpha_i, d_{\text{safe}}, \lambda$, risk weight $\gamma$, history window $N_H$
\State \textbf{Output:} Optimal control sequence $u_k$ and updated state $s_k$

\State \textbf{Step 1: Compute Predictive Matrices}
\State Compute stacked matrices $\mathcal{A}$ and $\mathcal{B}$ from system matrices $A,B$.

\State \textbf{Step 2: Initialize Historical Data (if first step)}
\If{$k = 0$}
\State Initialize $\bar{d}_i(0) = d_i(s_0)$ for all obstacles $i=1,\dots,N_{\text{obs}}$
\EndIf

\State \textbf{Step 3: Compute Evolutionary Risk Potential Field (ERPF)}
\For{$k=0,\dots, N-1$}
\For{each obstacle $i=1,\dots,N_{\text{obs}}$}
\State Calculate distance: $d_i(s_k)=\|Cs_k - p_i(k)\|$
\State Compute RPF value $\phi(d_i(s_k))$ using Eq. (\ref{eq:rpf})
\State Evaluate RPF risk: $V_{\text{RPF}}^i(s_k)=\alpha_i\,\phi(d_i(s_k))$
\State Update historical average distance $\bar{d}_i(k)$ using current and past distances
\State Compute evolution factor $\eta_i(k)$ using Eq. (\ref{eq:evolution_smooth})
\State Update ERPF risk: $V_{\text{ERPF}}^i(s_k)=\eta_i(k) V_{\text{RPF}}^i(s_k)$
\EndFor
\State Aggregate total ERPF risk: $V_{\text{ERPF}}(s_k)=\sum_{i=1}^{N_{\text{obs}}}V_{\text{ERPF}}^i(s_k)$
\EndFor

\State \textbf{Step 4: Formulate MPC Optimization Problem}
\State Define cost function J(u) in \ref{eq:cost}
\State Subject to system dynamics: $s_{k+1}=As_k+Bu_k$, with initial condition $s_0$

\State \textbf{Step 5: Solve MPC Optimization}
\State Solve the optimization problem: $\min_U J(U)$ with state and input constraints

\State \textbf{Step 6: Apply Control and Update State}
\State Apply the first optimal control input $u_0$ and update state: $s_0 \leftarrow s_1$

\State \textbf{Return:} Optimal control sequence $u_k$ and updated state $s_k$
\end{algorithmic}
\end{algorithm}

\section{Risk Quantification Based on Collision Ellipse Model}

The quantification of collision risk between vehicles in dynamic traffic environments requires mathematical models that capture both spatial and temporal dimensions of potential conflicts. This section presents an integrated approach to collision risk assessment using an adaptive ellipse model based on TTC and TWH parameters.

\subsection{Theoretical Foundation of Collision Ellipse}

The collision risk between two vehicles can be effectively represented by a dynamic elliptical region around the obstacle vehicle. This elliptical risk zone adapts its shape and size based on the kinematic relationship between the ego vehicle and the obstacle vehicle. The mathematical formulation of this adaptive ellipse begins with two primary temporal parameters: time-to-collision ($\text{TTC}$) and time window of hazard ($\text{TWH}$).

Time-to-collision represents the estimated time before two vehicles would collide if they maintained their current velocities and trajectories. Given the longitudinal position of the ego vehicle $x_{\text{ego}}$ and obstacle vehicle $x_{\text{obs}}$, along with their respective velocities $v_{\text{ego}}$ and $v_{\text{obs}}$, the $\text{TTC}$ can be calculated as:

\begin{equation}
\text{TTC} = \frac{x_{\text{obs}} - x_{\text{ego}}}{v_{\text{ego}} - v_{\text{obs}}}
\end{equation}

The time window of hazard ($\text{TWH}$) represents the duration during which a collision risk persists due to lateral position uncertainty. This parameter accounts for the temporal extension of risk beyond the specific $\text{TTC}$ instant, incorporating factors such as driver reaction time, maneuver execution uncertainty, and potential trajectory deviations.
\begin{figure}[t]
      \centering
      \includegraphics[width=\linewidth]{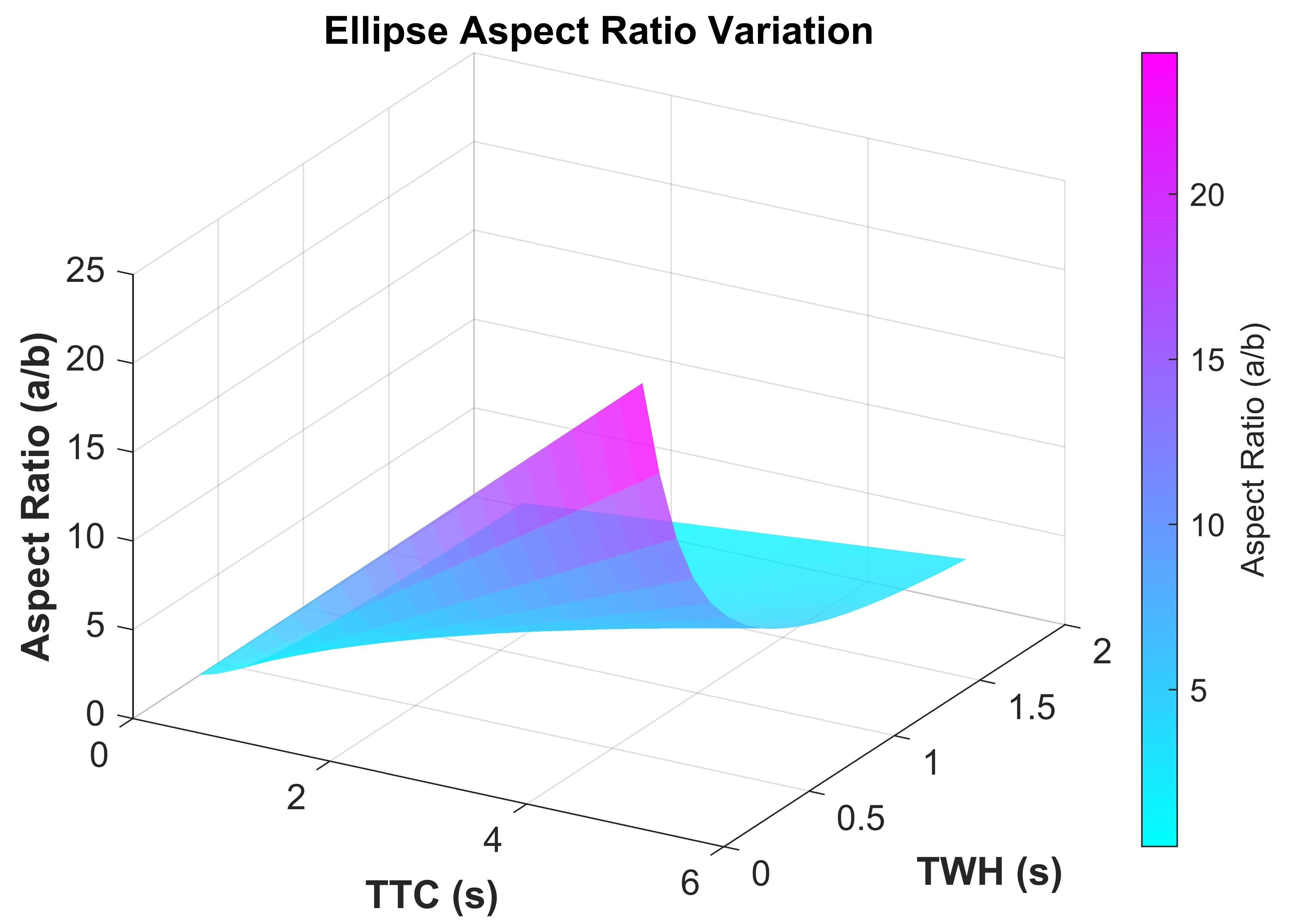}
      \caption{Evolution of Collision Ellipses with TTC and TWH.}
      \label{ttw}
\end{figure}
\subsection{Evolution Factor Ellipse Formulation}

The adaptive collision ellipse is mathematically defined by its semi-major axis $a$ and semi-minor axis $b$. These parameters are computed using the relative dynamics between vehicles and incorporating the temporal risk factors. To ensure that the dynamically generated collision ellipses remain within the vehicle's physically controllable region, we introduce kinematic constraints based on the maximum achievable longitudinal and lateral accelerations. Specifically, the semi-major axis is now defined as
\begin{equation}
a = \min\left( (v_{\text{ego}} - v_{\text{obs}}) \cdot \text{TTC},\; v_{\text{rel}} \cdot t_{\text{horizon}} + \frac{1}{2} a_{\max} t_{\text{horizon}}^2 \right),
\end{equation}
where $v_{\text{rel}} = |v_{\text{ego}} - v_{\text{obs}}|$, $a_{\max}$ is the maximum feasible deceleration, and $t_{\text{horizon}}$ denotes the planning horizon duration (e.g., $N\Delta t$ in MPC). This modification ensures that the longitudinal reach of the ellipse respects both perception and actuation limits.

Similarly, the lateral expansion is bounded to avoid unreasonably wide uncertainty regions. We redefine the semi-minor axis as

\begin{equation}
b = \sqrt{ \left( \frac{w_{\text{obs}}}{2} \right)^2 + \left[ \min\left( (v_{\text{ego}} - v_{\text{obs}}) \cdot \text{TWH},\; d_{\text{lat,max}} \right) \right]^2 },
\end{equation}
where $d_{\text{lat,max}}$ is the lateral motion budget determined by road geometry and control limitations (e.g., lane width), and we constrain $a \leq a_{\max} = 50$ m and $b \leq b_{\max} = 10$ m to ensure physical realizability. These constrained formulations mitigate the creation of unrealistically large ellipses that would otherwise overwhelm the MPC cost function and introduce false-positive risk cues.

where $w_{\text{obs}}$ represents the width of the obstacle vehicle. The ellipse axes thus adapt dynamically to the evolving traffic situation, with their ratio $a/b$ serving as an indicator of the ellipse eccentricity. 

Figure~\ref{ttw} has illustrated the ellipse variation among the TTC and TWH. It combines these effects into an aspect‐ratio surface. Tall, slender ridges occur when TTC is large but TWH is small, while broad, squat valleys appear when TWH dominates. This structure immediately distinguishes corridor‐like from diffuse hazard regimes. This volumetric view enables quick comparison of multiple scenarios, straightforward thresholding of unacceptable risk, and direct integration into real‐time control logic.

\begin{figure}[t]
      \centering
      \includegraphics[width=1\linewidth]{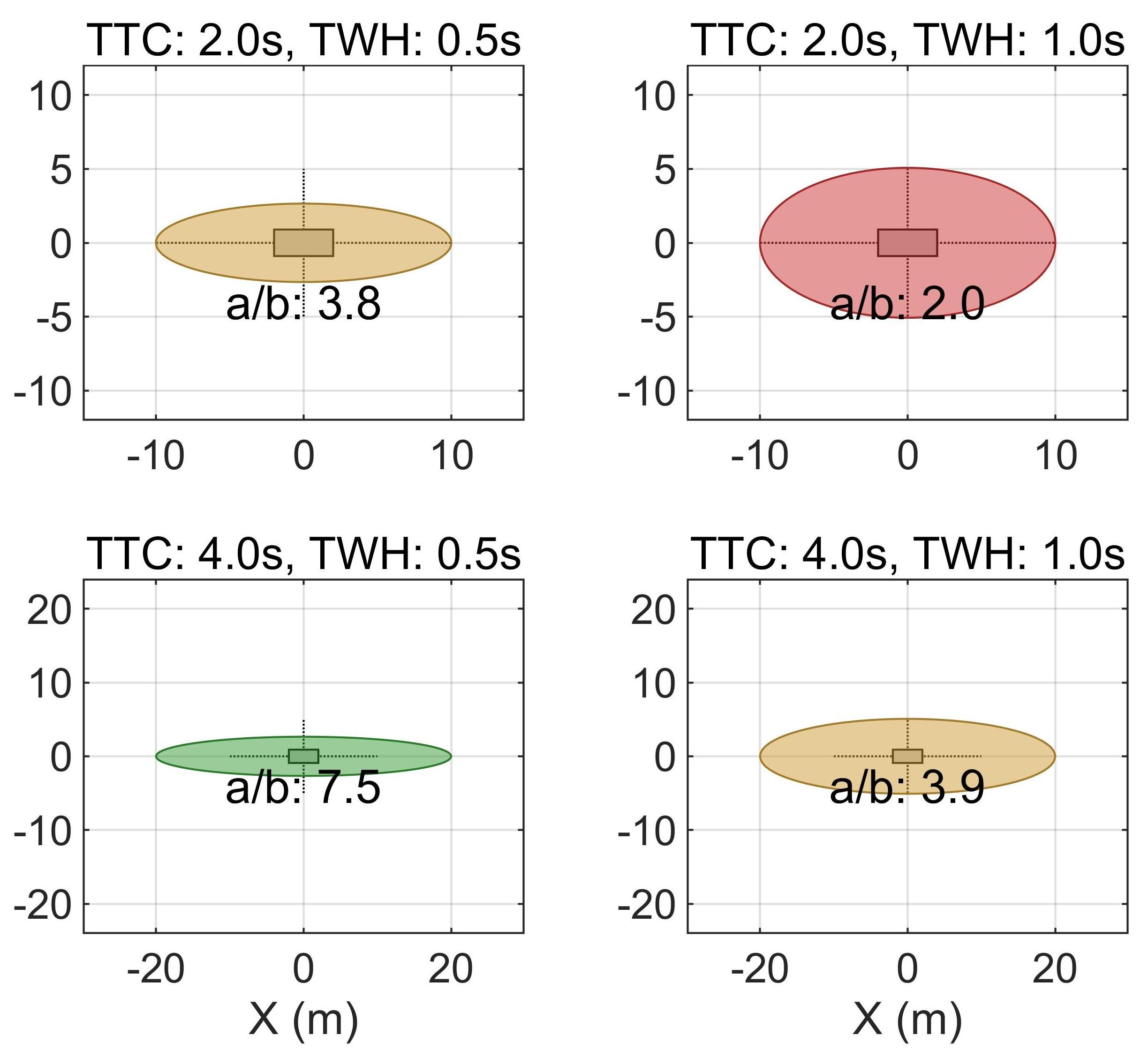}
      \caption{Collision ellipses for various combinations of TTC(rows: \(\mathrm{TTC}=2.0,4.0\)\,s) and TWH (columns: \(\mathrm{TWH}=0.5,1.0\)\,s).  Each subplot shows the obstacle (rectangle) and the overlaid risk ellipse with annotated aspect ratio \(a/b\).}
      \label{ttwcb}
\end{figure}

The 2×2 arrangement in Figure~\ref{ttwcb} systematically varies anticipation time against uncertainty window, revealing two complementary trends.  Moving down each column causes the ellipses to elongate dramatically along the longitudinal axis: anticipated collision distance grows roughly linearly with delay, producing very high aspect ratios (e.g.\ \(a/b\approx15\) at \(\mathrm{TTC}=2\,\mathrm{s},\;\mathrm{TWH}=1\,\mathrm{s}\)).  Conversely, advancing rightward in each row broadens the ellipses laterally, reducing aspect ratios and indicating that longer exposure windows yield more diffuse lateral spread.
In the top‐left corner (shortest TTC and smallest TWH), the ellipse is both shallow and moderately wide, reflecting imminent but spatially constrained hazard.  In the bottom‐right (longest TTC and largest TWH), the ellipse is both deep and wide, signifying a hazard that is both distant and highly uncertain in lateral position.  Intermediate grid cells expose mixed regimes—corridor‐like shapes when TTC dominates, and squat, circular‐leaning shapes when TWH dominates. Overlaying the numerical aspect‐ratio labels within each subplot provides an immediate, quantitative sense of "corridor versus cloud" risk: values above 5 indicate narrowly focused danger zones, while values below 1 mark almost isotropic uncertainty fields.  This visual and numerical pairing makes it trivial to pinpoint critical scenarios that demand different countermeasures, such as rapid deceleration for high‐aspect corridors, lateral avoidance for low‐aspect clouds.

\subsection{Evolution Factor for Risk Assessment}

The ellipse risk factor, denoted as $\text{ERF}$, serves as a normalized measure of collision risk based on the ellipse geometry.

\begin{equation}
\text{ERF} = \sqrt{\left(\frac{x}{a}\right)^2 + \left(\frac{y}{b}\right)^2}
\end{equation}

$\text{ERF} < 1$ indicates the ego vehicle is inside the risk ellipse, signifying high collision risk.
$\text{ERF} = 1$ indicates the ego vehicle is exactly on the ellipse boundary.
$\text{ERF} > 1$ indicates the ego vehicle is outside the risk ellipse, with larger values representing lower risk.

The ellipse risk factor can be further transformed into a risk metric $R$ through an exponential mapping:
\begin{equation}
R = e^{-\alpha \cdot (\text{ERF} - 1)}
\end{equation}
where $\alpha$ is a scaling coefficient that determines the risk decay rate as the evolution factor increases. This transformation bounds the risk between 0 and 1, with 1 representing maximum risk when $\text{ERF} \leq 1$ and asymptotically approaching 0 as the distance from the ellipse increases.

\subsection{Parameter Sensitivity Analysis}

The adaptive nature of the collision ellipse model is fundamentally governed by the relationship between $\text{TTC}$ and $\text{TWH}$ parameters. A comprehensive analysis reveals distinct patterns in ellipse morphology across different parameter combinations.

For small $\text{TTC}$ values (e.g., $\text{TTC} < 1.0$ s), the collision ellipse exhibits a more circular shape, as the semi-major and semi-minor axes become comparable in magnitude. This represents imminent collision scenarios where the risk is concentrated in close proximity to the obstacle vehicle. The aspect ratio $a/b$ approaches unity as $\text{TTC}$ decreases, particularly when combined with larger $\text{TWH}$ values.

As $\text{TTC}$ increases (e.g., $\text{TTC} > 2.0$ s), the ellipse becomes progressively elongated along the direction of travel, with the aspect ratio $a/b$ increasing substantially. For $\text{TTC} = 4.0$ s combined with small $\text{TWH}$ values (e.g., $\text{TWH} = 0.2$ s), the aspect ratio can exceed 5.0, creating a highly eccentric ellipse that extends far ahead of the obstacle vehicle.

The $\text{TWH}$ parameter primarily influences the lateral expansion of the risk ellipse. Larger $\text{TWH}$ values (e.g., $\text{TWH} > 1.0$ s) generate wider ellipses that account for greater lateral uncertainty in vehicle trajectories. This lateral expansion becomes particularly significant at higher relative velocities, as the product $(v_{\text{ego}} - v_{\text{obs}}) \cdot \text{TWH}$ increasingly dominates the semi-minor axis calculation.

\section*{Benefit of Sigmoid-based Evolution Factor in Optimization}
Note: This section discusses the evolution factor $\eta(k)$ used in the ERPF framework (Eq. \ref{eq:evolution_smooth}), which is distinct from the ellipse risk factor (ERF) introduced in the collision ellipse model.
Let $J(U)$ be the ERPF-MPC cost function defined as
\[
J(U) = \sum_{k=0}^{N-1} \left( \|s_k - s_k^{\mathrm{ref}}\|_Q^2 + \|u_k\|_R^2 + \gamma \cdot \eta(k) \cdot V_\text{RPF}(s_k) \right),
\]
where $\eta(k)$ modulates risk amplification. 

If $\eta(k)$ is a binary indicator function, i.e.,
\[
\eta(k) = 1 + \lambda \cdot \mathbb{I}\{\bar{d}(k) < d(s_k)\},
\]
then $J(U)$ becomes non-smooth and piecewise-defined. This introduces nondifferentiability at the boundary $d(s_k) = \bar{d}(k)$, degrading solver stability in gradient-based optimization.

By replacing $\eta(k)$ with a sigmoid-based formulation,
\[
\eta(k) = 1 + \lambda \cdot \sigma\left(\frac{\bar{d}(k) - d(s_k)}{d_{\text{safe}}}\right),
\]
where $\sigma(z) = 1 / (1 + e^{-z})$, we ensure that:
\begin{itemize}
    \item $\eta(k)$ is smooth, bounded, and differentiable;
    \item The overall cost function $J(U)$ remains continuously differentiable over the planning horizon;
    \item The optimizer receives earlier, graded feedback when $\bar{d}(k) > d(s_k)$, avoiding abrupt trajectory shifts.
\end{itemize}

This smooth evolution improves convergence, ensures better safety margin anticipation, and enhances trajectory smoothness.

%%%%%%%%%%%%%%%%%%%%%%%%%%%%%%%%%%%%%%%%%%%%%%%%%%%%%%%%%%%%%%%%%%%%%%%%%%%%%%%%%第七节

\section{Simulation Results}
\label{sec4}

The simulations were designed to verify the safety, stability and efficiency of the proposed method. The simulations were conducted on a computer with the Ubuntu 18.04.6 LTS OS, a 12th generation 16-thread Intel\textsuperscript{\textregistered}Core\texttrademark\ i5-12600KF CPU, an NVIDIA GeForce RTX 3070Ti GPU, and 16GB of RAM. The simulation results are obtained in MATLAB R2024b. 

To verify the effectiveness of our proposed ERPF-MPC, we constructed a series of experimental scenarios. To demonstrate that ERPF can better alert the AV to risky areas without affecting normal driving compared to RPF, we compared the field strength during the vehicle's approach and stable driving for both ERPF and RPF. Additionally, we used two scenarios, namely lane changing and overtaking, to evaluate the driving generalization capability of the proposed ERPF-MPC. Each scenario involves dense HDVs surrounding the AV, increasing the risk of collisions and the difficulty of driving without collisions. Finally, to highlight the superior performance of the proposed ERPF, we conducted comparisons with other popular benchmark algorithms under these two scenarios.

\begin{figure}[t]
    \centering
    \includegraphics[width=0.5\textwidth]{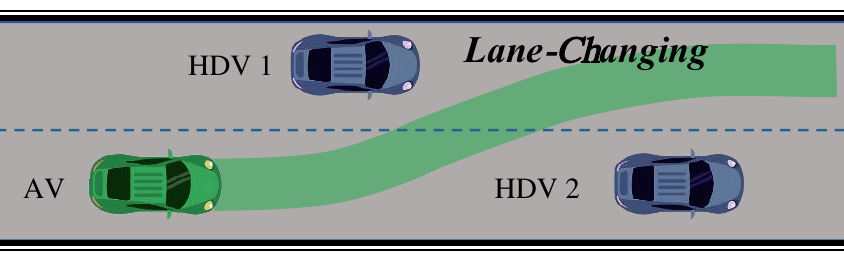}
    \caption{Initial positions for the AV and HDVs in the considered lane-changing scenario 1.}
    \label{n2}
\end{figure}

\begin{figure}[ht]
    \centering
    \includegraphics[width=0.4\textwidth]{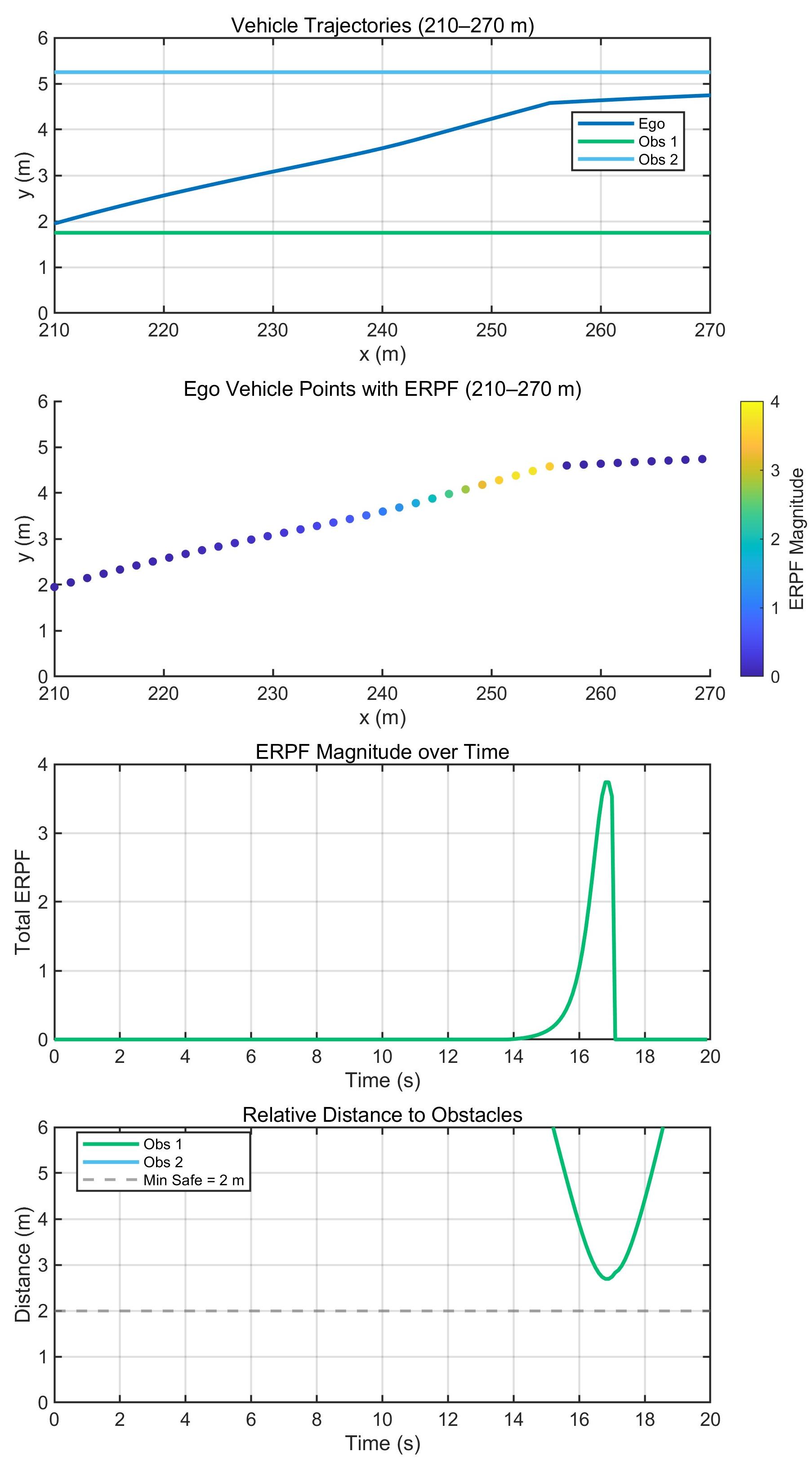}
    \caption{EPRF–guided safe lane change Under multi-vehicle interaction in considered lane-changing scenario 1}
    \label{bpp}
\end{figure}
Figure~\ref{n2} shows the ERPF simulation setup:
an AV starts at $(0,\,1.75)$ m with speed $30\,$m/s;
HDV~1 is at $(50,\,1.75)$ m with $12\,$m/s;
and HDV~2 is at $(40,\,5.25)$ m with $20\,$m/s.
The dashed lines mark the centers of lane~1 ($y=1.75\,$m)
and lane~2 ($y=5.25\,$m), with the shaded region indicating the drivable area.

Figure~\ref{bpp} considers three vehicles with the following start positions: the ego‐vehicle starts at $(x,y)=(0,1.75)\,$m and drifts upward between 50–100 m to avoid two obstacles: Obs 1 in the same lane starting at $(50,1.75)\,$m and Obs 2 in the opposite lane starting at $(40,5.25)\,$m. The two middle sub-figures show total ERPF repulsion, which spikes near 15 s and 255 m reaching around 4 to trigger the maneuver. The bottom sub-figure plots the ego vehicle’s distance to each obstacle. The blue curve reaches a minimum of approximately 2.3 meters, which stays just above the 2-meter safety threshold indicated by the dashed line. In contrast, the orange curve remains at a consistently safer distance throughout.

\begin{figure}[t]
    \centering
    \includegraphics[width=0.5\textwidth]{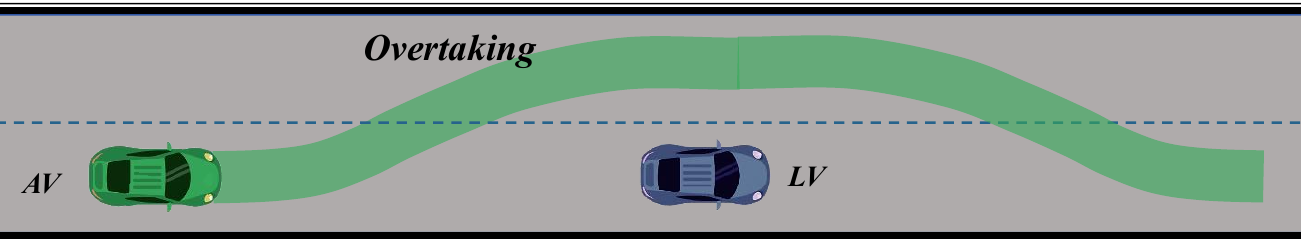}
    \caption{Initial positions for the AV and HDVs in the considered Overtaking scenario 2.}
    \label{n22}
\end{figure}

Figure~\ref{n22} shows the ERPF simulation of overtaking where there is a LV in front of the AV. During the overtaking, the AV is required to avoid any collisions with the LV.

Figure~\ref{bpp} illustrates a comparative analysis of planned trajectories under three different planning strategies: Static Risk Potential Field (RPF), Original Elliptical Risk Potential Field (ERPF), and the proposed Improved ERPF (ERPF-Improved). The left panel shows the spatial evolution of lateral positions against longitudinal distances, while the right subplots provide temporal dynamics of lateral position and inter-vehicle distance.

Notably, the ERPF-Improved strategy demonstrates a clear overtaking success that neither the Static RPF nor the Original ERPF manages to achieve. As seen in the main trajectory plot, ERPF-Improved (blue curve) successfully navigates across the lane boundary, completing a full lane change and merging back safely after overtaking the leading vehicle (LV). In contrast, the Static RPF (red curve) hesitates mid-course, exhibiting excessive lateral fluctuation and failing to commit to a full overtake. Similarly, the Original ERPF (magenta dashed curve) initiates the lane change but ultimately stabilizes without completing the maneuver, remaining trapped behind the LV.
\begin{figure}[ht]
    \centering
    \includegraphics[width=0.4\textwidth]{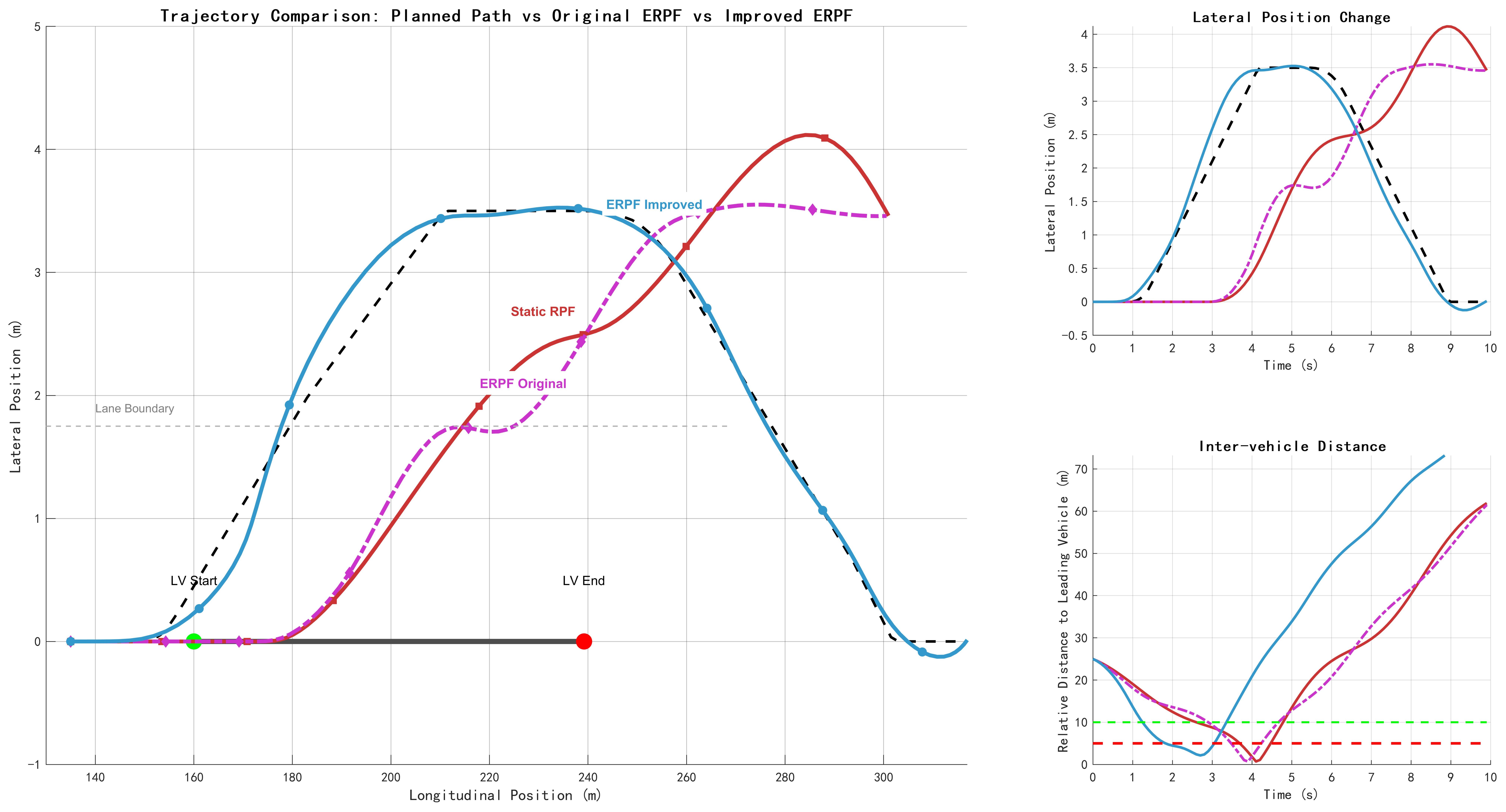}
    \caption{EPRF-MPC–guided (EPRF-Improved) safe overtaking compared to RPF, and ERPF (EPRF-Original) , in considered overtaking scenario 2}
    \label{bpp}
\end{figure}
The lateral position over time further highlights this superiority: ERPF-Improved rapidly reaches the target lane around $t=3.5$ seconds and maintains lateral stability afterward. Conversely, both baseline methods show delayed or incomplete transitions. The inter-vehicle distance plot confirms that only ERPF-Improved consistently increases separation after the critical interaction period, maintaining safe margins above the safety thresholds (green and red dashed lines). The other methods experience temporary violations of safe following distance due to hesitation or incomplete maneuvers.

\begin{figure}[ht]
    \centering
    \includegraphics[width=0.4\textwidth]{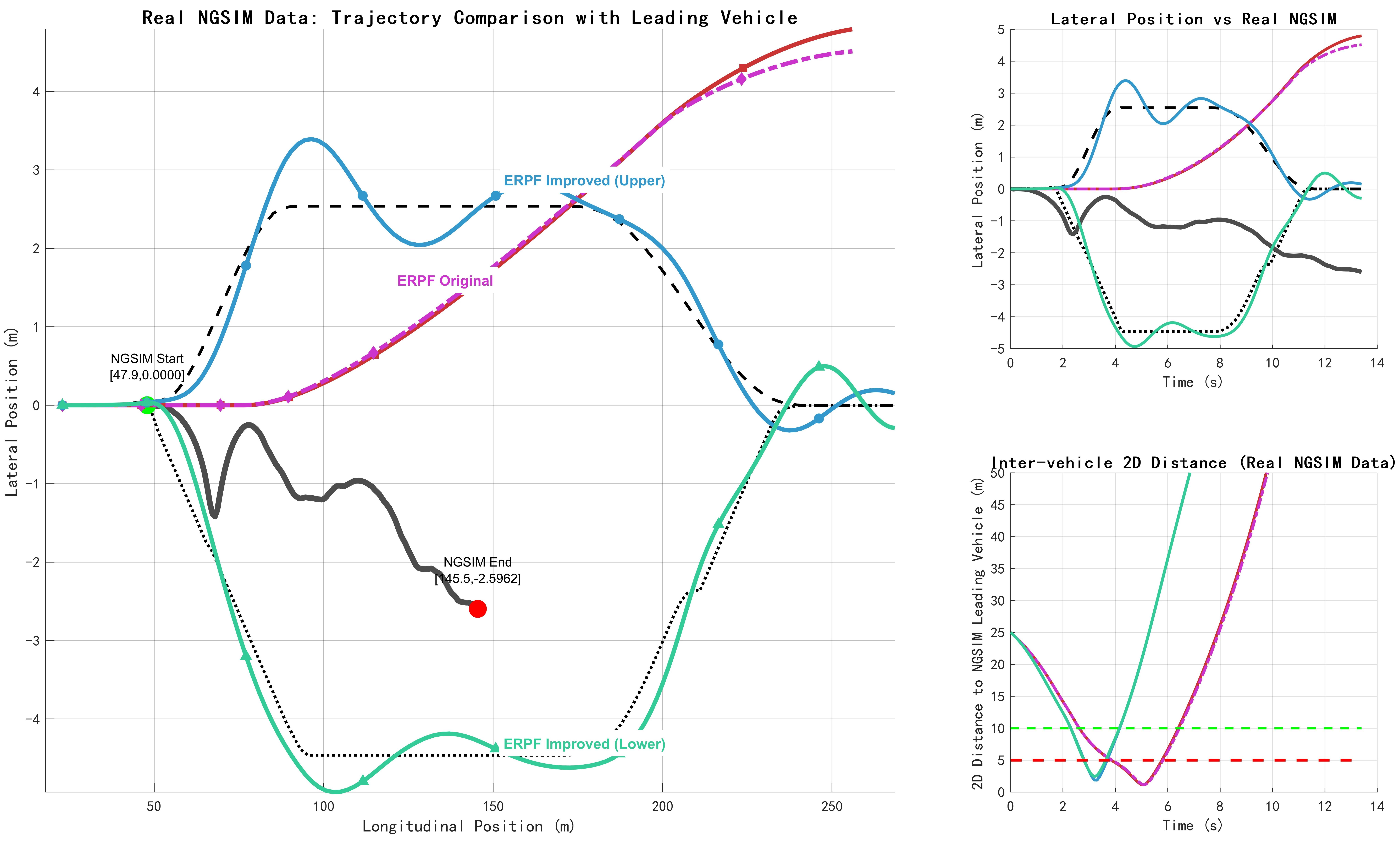}
    \caption{EPRF-MPC–guided (EPRF-Improved) safe overtaking compared to RPF, and ERPF (EPRF-Original) , in considered overtaking scenario using NGSIM real-world data}
    \label{bppp}
\end{figure}

To assess the practical feasibility of our proposed ERPF-Improved planner, we conduct trajectory evaluations using real-world vehicle behavior from the NGSIM dataset. As shown in Figure~\ref{bppp}, the black solid line represents the actual recorded trajectory of the LV, and our method is evaluated against this ground truth. 

The ERPF-Original planner (magenta dashed) maintains a consistent rightward lane shift, with minimal adaptation to the real traffic flow, eventually diverging from the real LV. In contrast, our ERPF-Improved planner explores both upper (blue) and lower (green) overtaking maneuvers. These dual strategies demonstrate the model’s flexibility in adapting to different environmental constraints.

Notably, the lower overtaking path exhibits a more cautious and conservative behavior, characterized by a longer planning distance and more gradual lateral deviations. This aligns well with real-world driving logic, where downward overtaking often requires larger safety margins and greater anticipation due to visual occlusion and limited acceleration space. The upper path, while more direct, reflects a higher risk strategy that may be less appropriate in dense traffic.

Temporal plots on the right reinforce these insights. The lateral position over time shows that ERPF-Improved (Lower) reaches greater lateral extents earlier and maintains smoother transitions compared to the other planners. The 2D inter-vehicle distance plot further confirms that the ERPF-Improved (Lower) path ensures consistently safer separation from the real NGSIM leading vehicle, exceeding the predefined safety thresholds (dashed green/red lines) throughout the maneuver.

\subsection{Benchmark Comparisons for Performance Evaluation}
\begin{figure*}[htbp]
    \centering
    \includegraphics[width=0.7\textwidth]{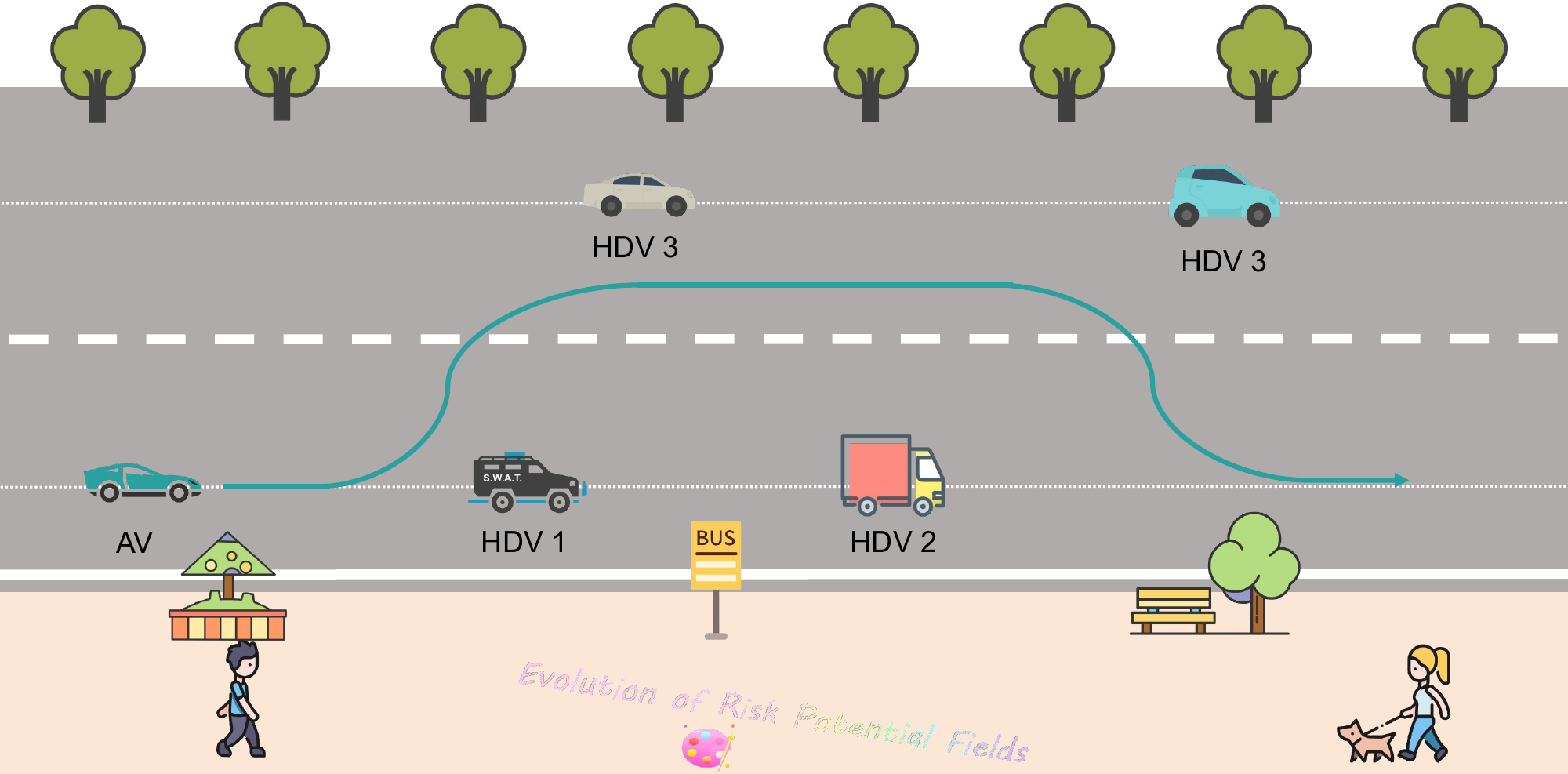}
    \caption{Overtaking scenario 2 diagram for an AV and surrounding HDVs in a smart city.}
    \label{fig:lcs}
\end{figure*}

Figure~\ref{fig:lcs} illustrates a two‐lane highway scenario: an autonomous vehicle (AV, teal) travels at 35 m/s in the inner lane behind two slower HDVs, both 15 m/s at 50 m and 80 m respectively. In the outer lane, HDV 3 moves at 30 m/s at 40 m, and HDV 4 at 15 m/s at 100 m. The cyan trajectory shows the AV merging into the outer lane to overtake HDV 1 and HDV 2, then re‐merging into the inner lane ahead of them. Background elements provide roadside context without affecting the vehicle interactions.

\begin{figure}[ht]
    \centering
    \includegraphics[width=0.4\textwidth]{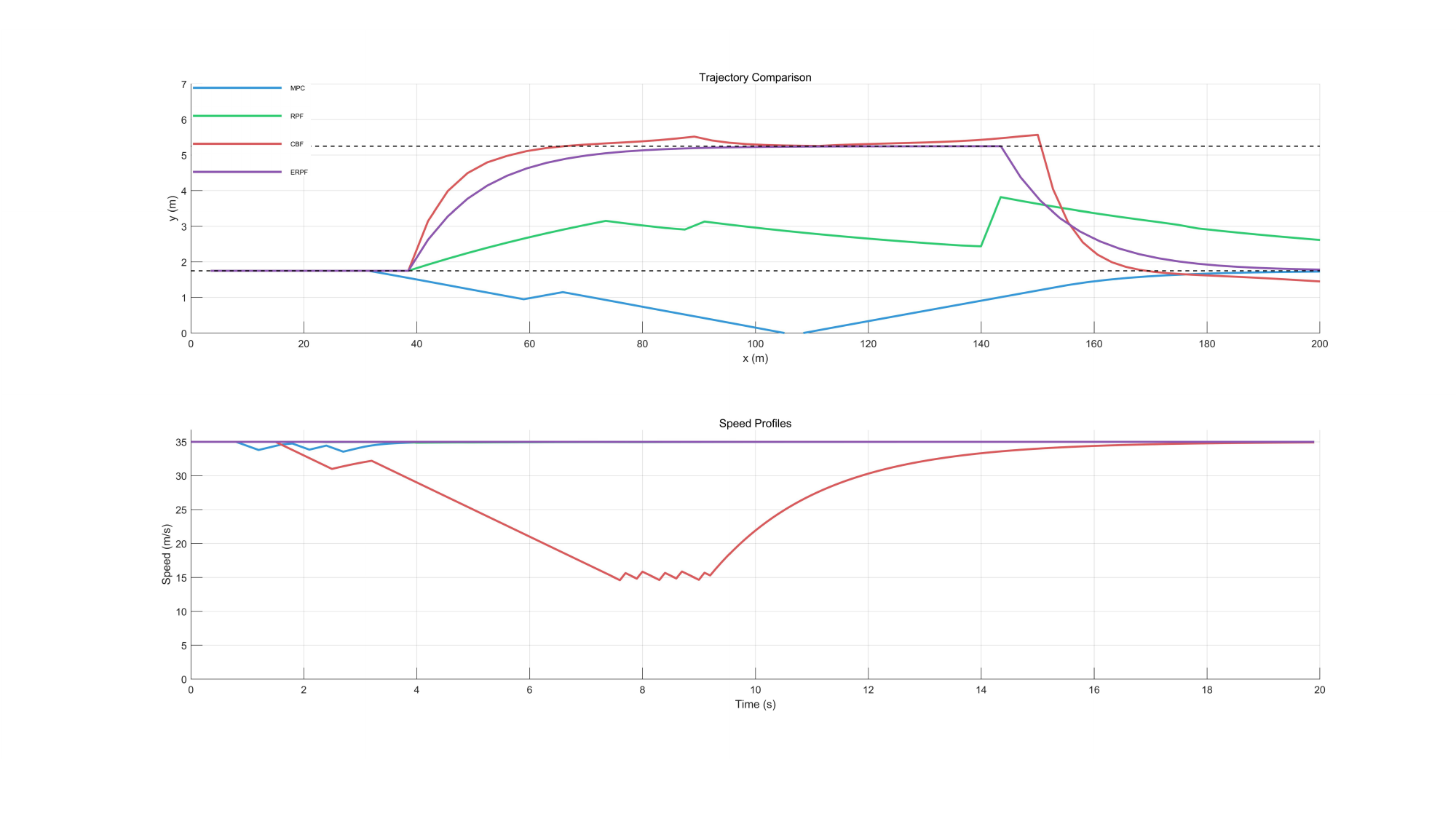}
    \caption{Trajectory and speed comparison of MPC, RPF, CBF and ERPF controllers in the considered overtaking scenario 2.}
    \label{fig:overcom}
\end{figure}
Figure~\ref{fig:overcom} upper-side shows the vehicle’s lateral position \(y\) versus longitudinal distance \(x\) under four control schemes: MPC (blue), RPF (green), CBF (red) and ERPF (purple).  The dashed lines mark the centers of lane 1 in \(y = 1.75\) m and lane 2 in \(y = 5.25\) m.  MPC stays in lane 1 throughout; RPF executes a shallow lane change around \(x\approx40\) m but remains in lane 2 for only part of the maneuver; CBF and ERPF both perform robust lane changes into lane 2, reaching \(y\approx5.2\) m by \(x\approx60\) m, then return to lane 1 around \(x\approx150\) m, with ERPF’s curve slightly smoother on exit.  

Figure~\ref{fig:overcom} bottom-side plots the speed profiles over time.  ERPF maintains the reference speed of 35 m/s nearly flat, MPC and RPF exhibit mild speed fluctuations around 30–35 m/s, while CBF decelerates sharply to about 15 m/s during the lane change  before accelerating back to 35 m/s by second 16. This comparison highlights the trade‐offs between predictive performance, safety interventions, and smoothness across the four controllers.
\begin{figure}[htbp]
    \centering
    \includegraphics[width=0.4\textwidth]{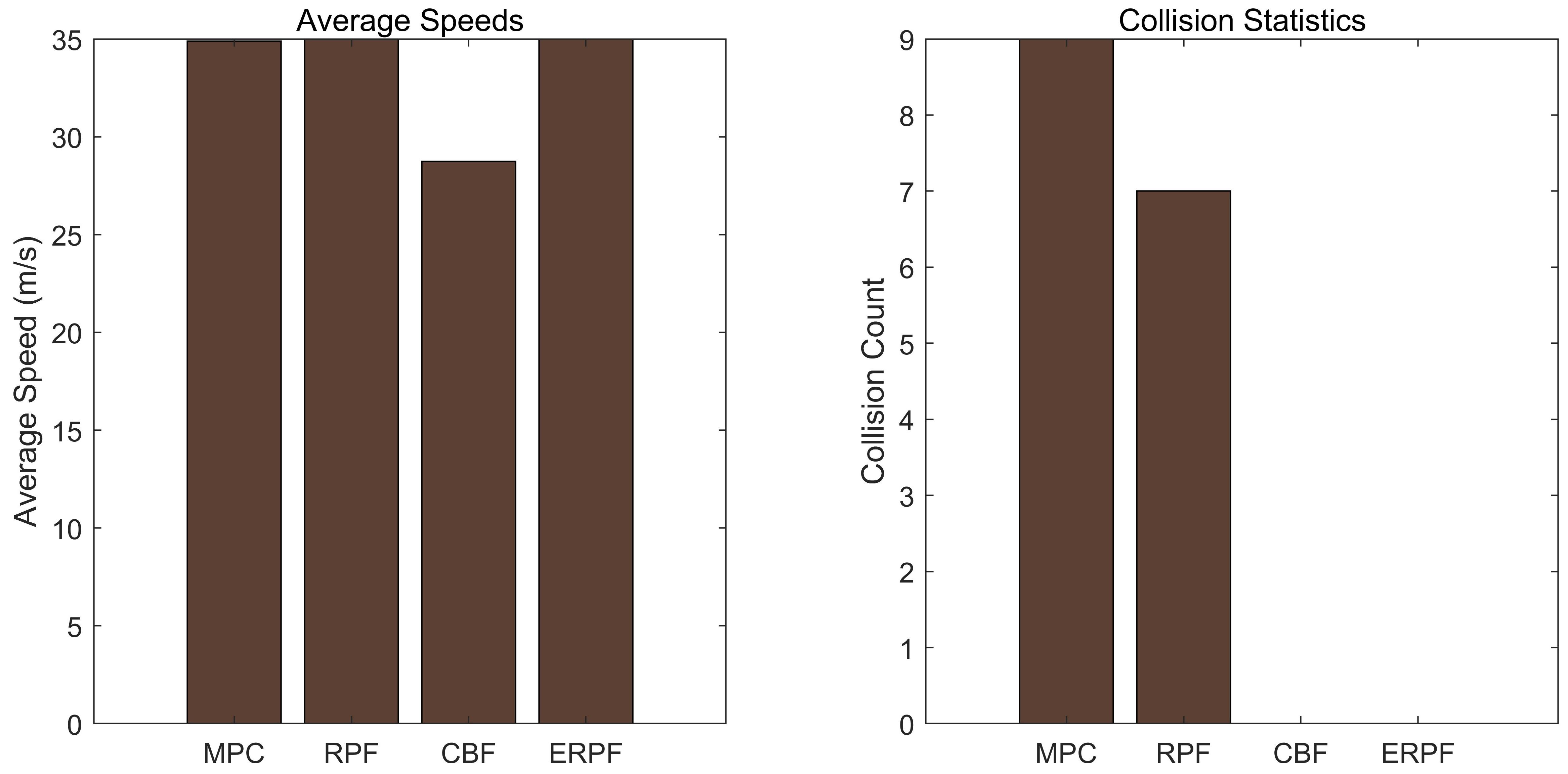}
    \caption{Average speeds and collision counts for MPC, RPF, CBF, and ERPF.}
    \label{fig:overtaking_trajectories}
\end{figure}

\begin{figure*}[htbp]
  \centering
  \includegraphics[width=0.7\textwidth]{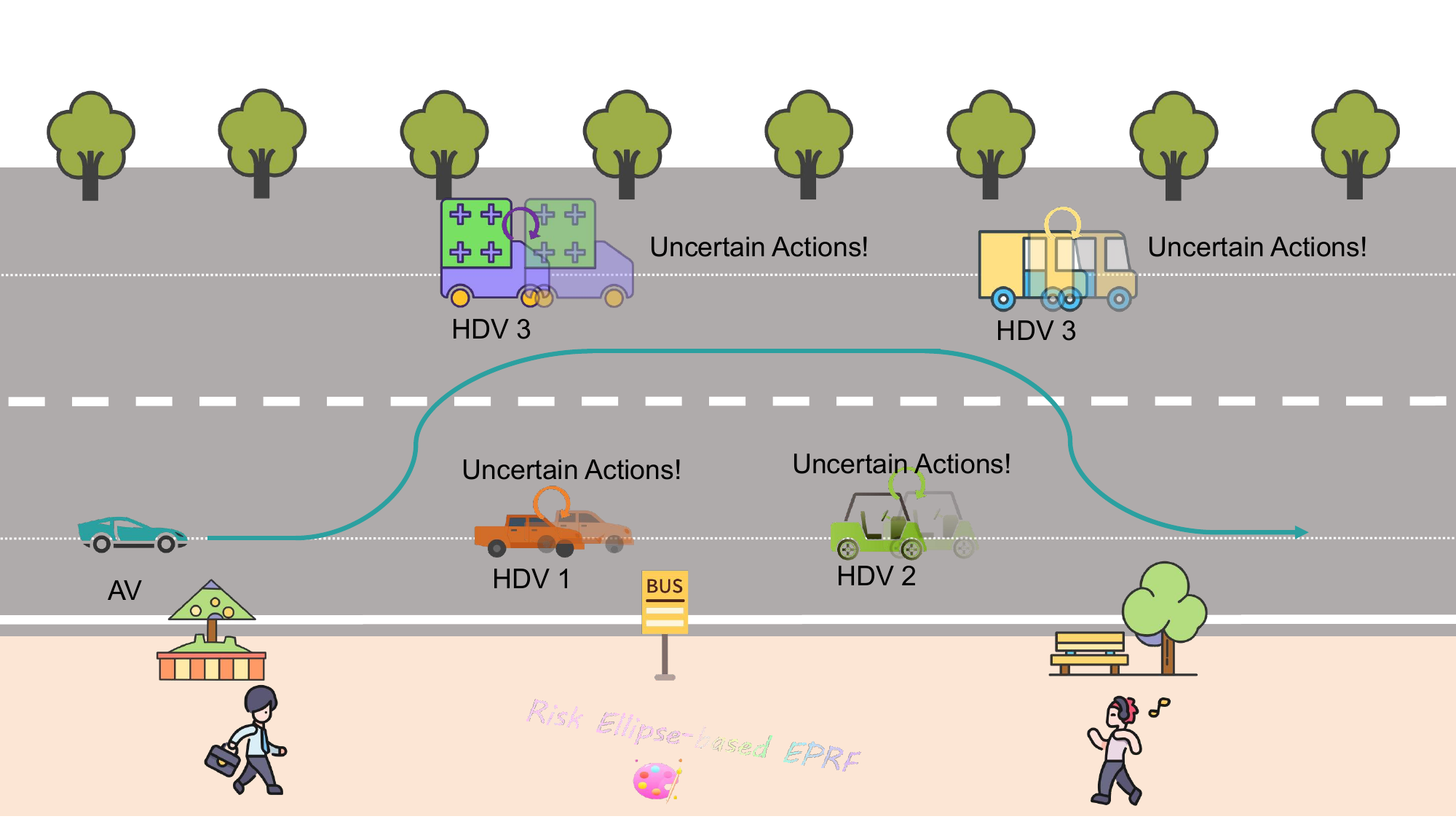}
  \caption{High‐level scenario: an AV navigating between four HDVs, each annotated with a semi‐transparent “ghost” and a rotation arrow indicating uncertain acceleration maneuvers.}
  \label{fig:scenario}
\end{figure*}
\begin{figure}[ht]
  \centering
  \includegraphics[width=0.4\textwidth]{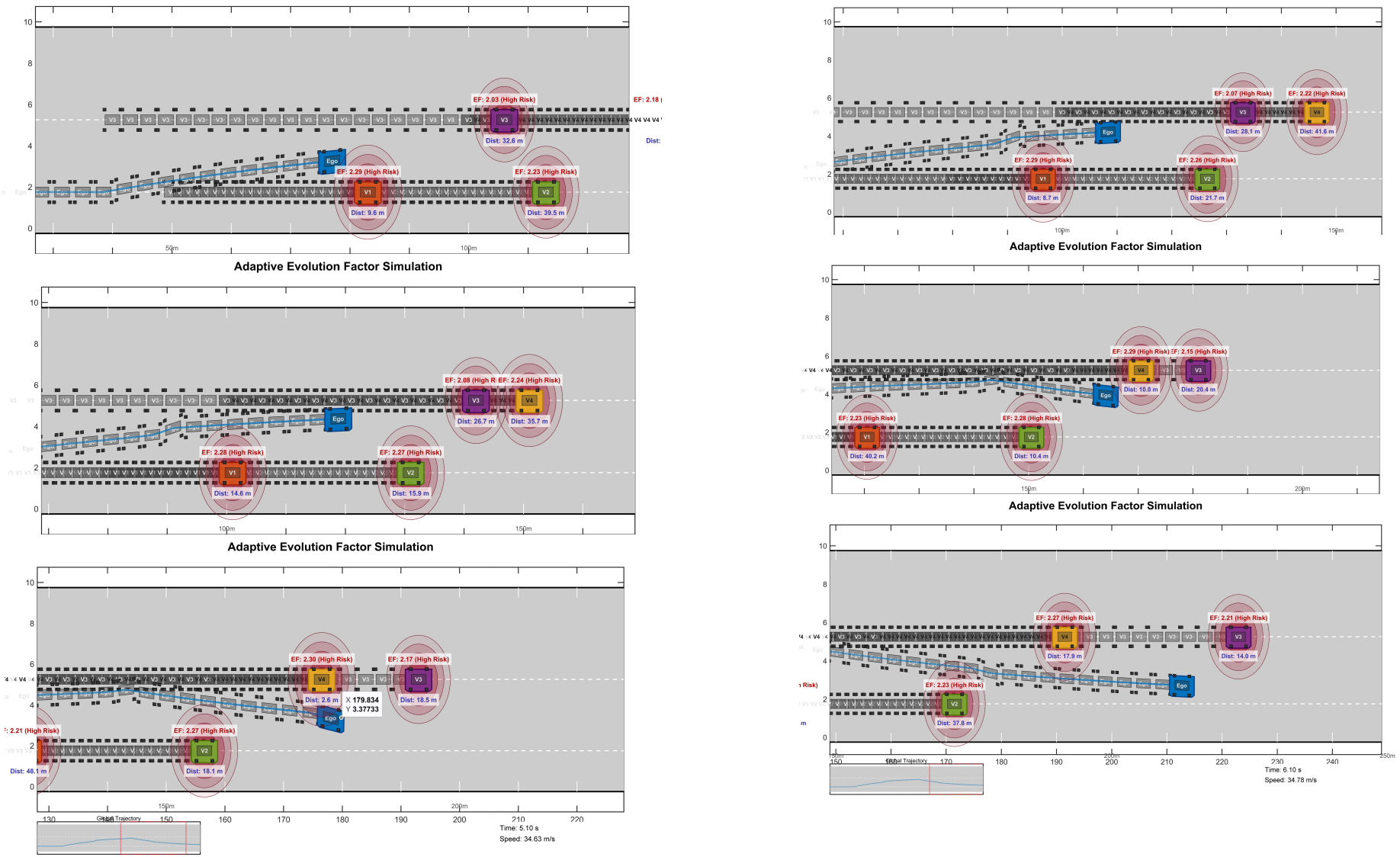}
  \caption{Snapshots from the adaptive evolution‐factor simulation showing the ego vehicle (blue), three human‐driven vehicles (V1–V4), and their dynamically computed risk ellipses labeled with EF and inter-vehicular distance.}
  \label{fig:adaptive_sim}
\end{figure}

\begin{figure}[ht]
  \centering
  \includegraphics[width=0.3\textwidth]{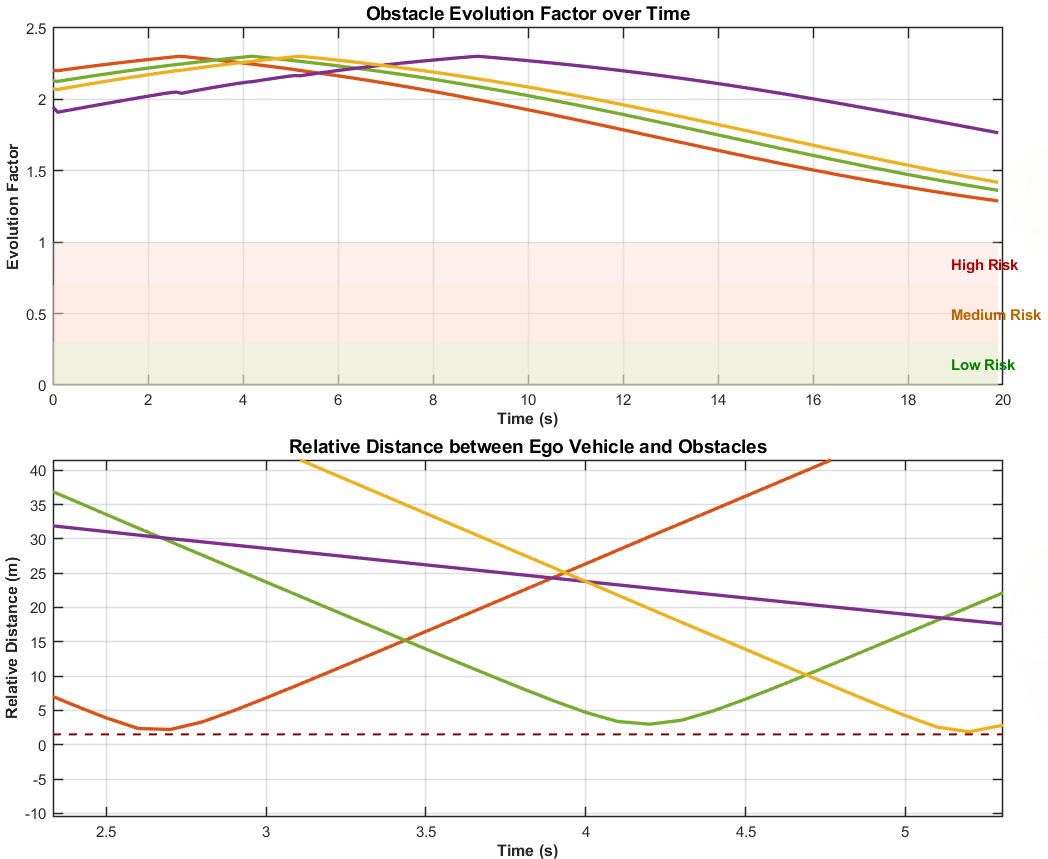}
  \caption{Top: EF for each obstacle over time. Bottom: corresponding relative distances between the ego vehicle and each obstacle, with the red dashed line marking the minimum safe following distance.}
  \label{fig:ef_distance}
\end{figure}

\begin{figure}[h]
\centering
\includegraphics[width=0.2\textwidth]{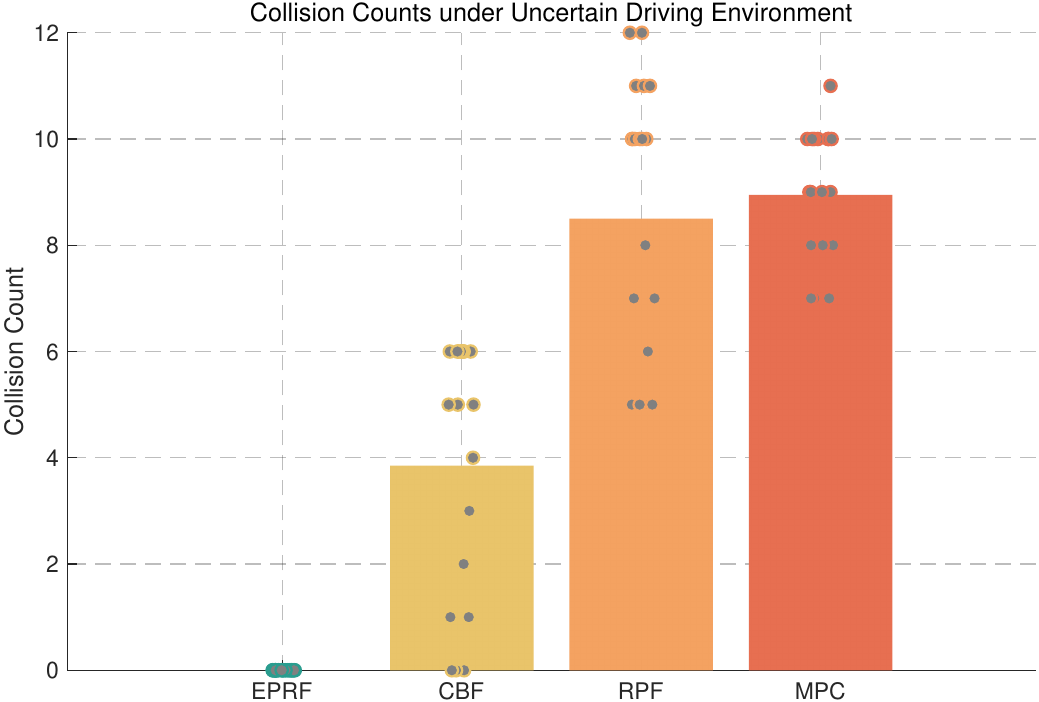}
\caption{Collision counts under uncertainty: EPRF vs.\ CBF vs.\ RPF vs.\ MPC (20 runs each).}
\label{fig:collision_counts}
\end{figure}
Figure~\ref{fig:overtaking_trajectories} compares the four controllers on two key metrics.  In the Average Speeds plot, MPC, RPF, and ERPF all achieve near–reference speeds around 35 m/s, whereas CBF falls back to about 29 m/s.  In the Collision Statistics plot, both CBF and ERPF incur zero collisions, while MPC records 9 and RPF 7 collisions. Thus ERPF combines the high throughput of MPC and RPF, with the perfect safety record of CBF, demonstrating superior overall performance.

Figure~\ref{fig:scenario} depicts an AV initially at $(x,y)=(5,1.75)\,\mathrm{m}$ with speed $45\,\mathrm{m/s}$ navigating among three human‐driven vehicles—HDV 1 at $(55,1.75)\,\mathrm{m}$, $15\,\mathrm{m/s}$; HDV 2 at $(45,5.25)\,\mathrm{m}$, $30\,\mathrm{m/s}$; and HDV 3 at $(85,1.75)\,\mathrm{m}$, $15\,\mathrm{m/s}$.  Each HDV is shown with a semi‐transparent “ghost” overlay and a curved arrow to indicate uncertain future accelerations, signaling that their TTC and TWH estimates will yield risk ellipses whose semi‐major and semi‐minor axes dynamically stretch and widen.  The AV’s teal trajectory weaves between these evolving envelopes, maintaining safe separation by anticipating both longitudinal reach and lateral dispersion of the HDVs’ possible maneuvers.  A shaded sidewalk with pedestrian and street‐furniture icons further constrains lateral deviation, emphasizing that the risk‐ellipse framework must respect off‐road safety margins.  Altogether, this single‐frame illustration demonstrates how uncertain human actions are algebraically absorbed into spatio–temporal risk fields that drive the AV’s adaptive lane‐keeping and collision‐avoidance strategy.

Figure~\ref{fig:adaptive_sim} presents snapshots from the adaptive evolution factor simulation. At each control tick, the system computes the TTC and TWH for each HDV, based on their instantaneous relative distance and velocity. These values are then mapped into EF, which scale the semi-major and semi-minor axes of the risk ellipses. When an HDV decelerates or accelerates unexpectedly, as indicated by the circular uncertainty arrows, the corresponding ellipse expands forward or laterally to account for the worst-case trajectories. During steady straight-line motion, narrow ellipses allow the AV to maintain a comfortable headway. When an HDV's ellipse exceeds a predefined safety threshold, such as when EF surpasses 2.2 or the ellipse intersects the AV’s projected path, the controller combines longitudinal deceleration with subtle lateral adjustments. These actions include applying small braking pulses to gradually reduce speed and shifting slightly within the lane to prepare for a possible lane change.
\begin{figure}[t]
    \centering
    \includegraphics[width=0.4\textwidth]{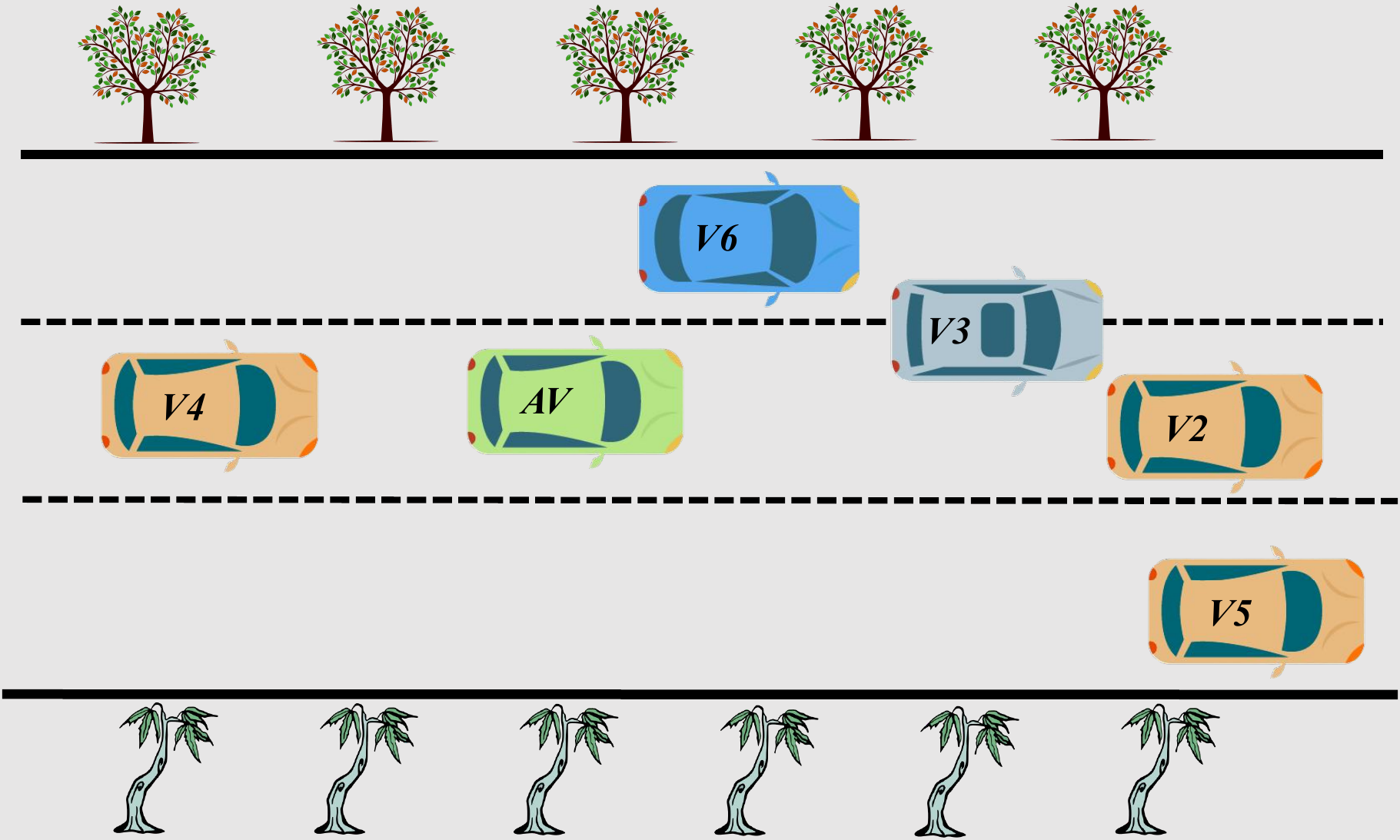}
    \caption{Extreme-High‐level scenario: an AV non-target navigating between five HDVs, in a three-lane highway.}
    \label{ncp}
\end{figure}

\begin{figure}[htbp]
    \centering
    \includegraphics[width=1\linewidth]{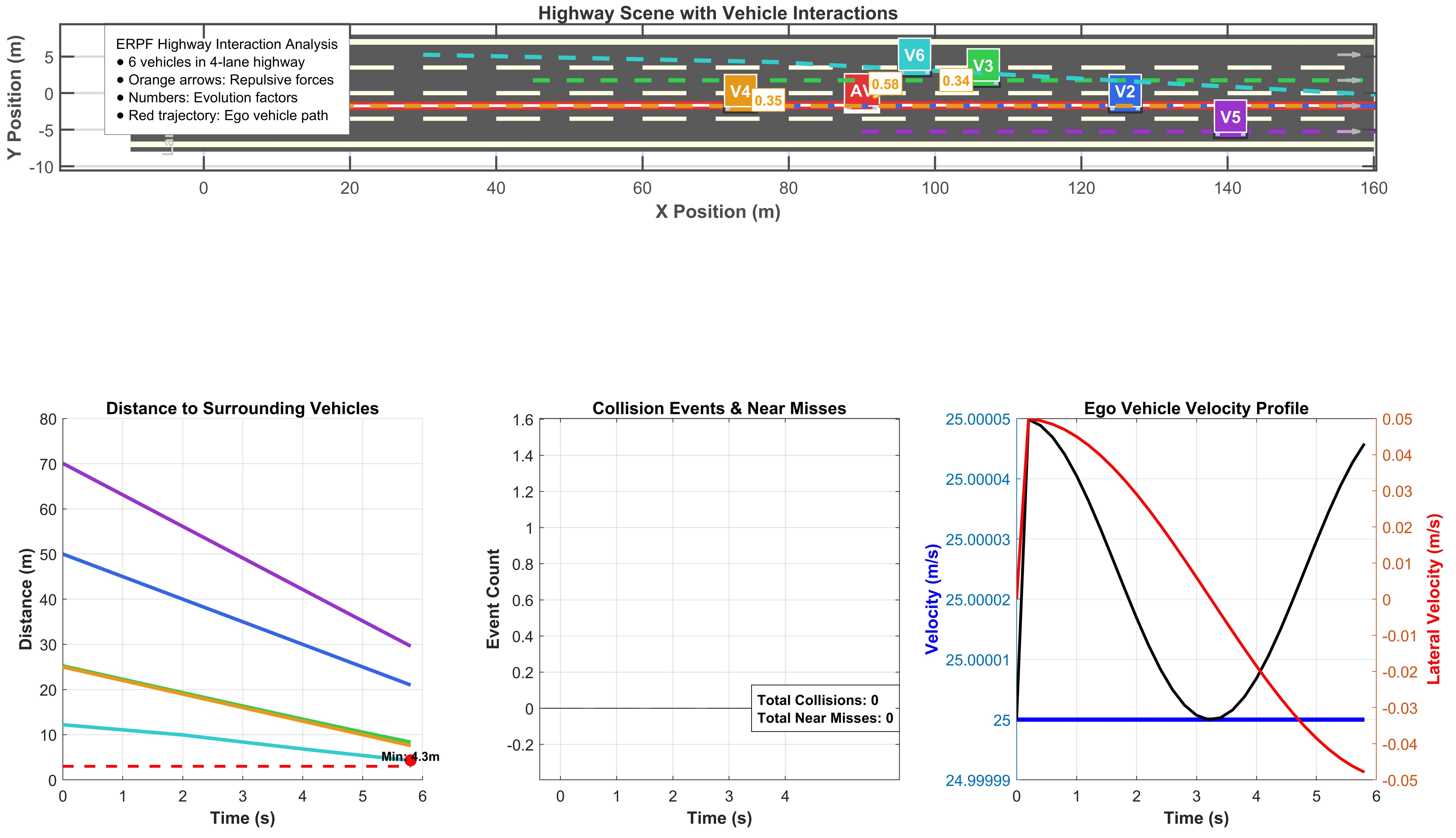}
    \caption{ERPF-based interaction analysis in a 3-lane highway with six vehicles. Repulsive forces, evolution factors, distances, and collision statistics are visualized.}
    \label{fig:highway_safety}
\end{figure}

Throughout the maneuver, the AV continuously replans its projected trajectory in real time, treating the evolving ellipses as moving obstacles in the path planner. If two adjacent ellipses begin to overlap, the planner prioritizes the tighter corridor and initiates an early and smooth lane transition before the high-risk region fully develops. As the ellipses shrink due to stabilized HDV motion, the controller relaxes braking and recenters the vehicle within the lane. This closed feedback loop, illustrated in Figure~\ref{fig:adaptive_sim}, ensures that every uncertain human action is immediately reflected in the geometry of the risk ellipses and translated into smooth and safe control responses.

Figure~\ref{fig:ef_distance} illustrates the adaptive inflation of risk ellipses based on obstacle proximity. The top panel shows EF values over time for different obstacles, with color-coded threat levels: green, orange, and red. When obstacles approach or decelerate, EF curves rise into higher risk regions, expanding the corresponding collision ellipses. The bottom panel displays relative distances, where EF peaks typically occur before potential safety violations, demonstrating predictive capability. This synchronized behavior creates an intuitive geometric safety buffer that inflates proactively and deflates as threats subside.

\begin{figure}[t]
    \centering
    \includegraphics[width=\linewidth]{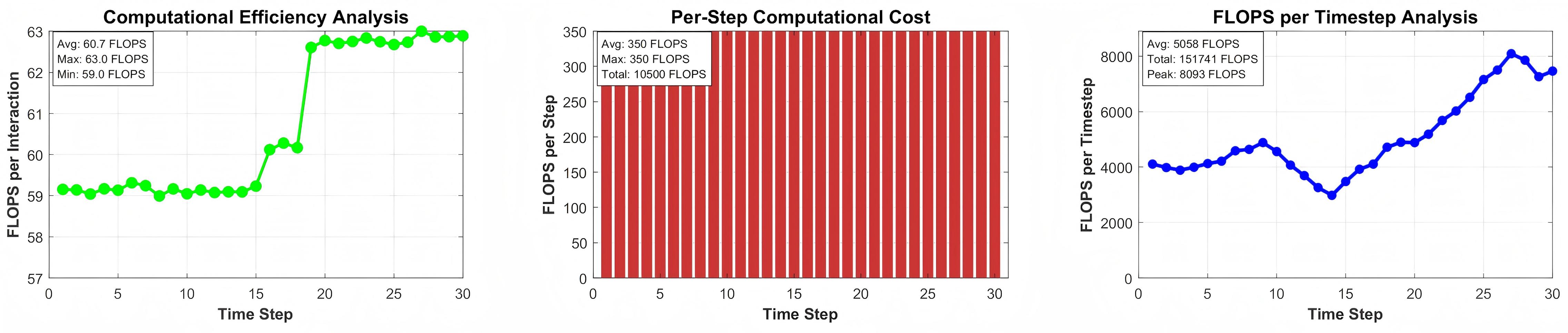}
    \caption{Computational analysis of the proposed ERPF planner: per-interaction efficiency, per-step cost, and overall per-timestep FLOPs.}
    \label{fig:runtime}
\end{figure}

Figure \ref{fig:collision_counts} demonstrates superior safety performance across four control schemes over 20 trials. ERPF achieved zero collisions in all runs, while CBF averaged 3 collisions (range 0-6), RPF averaged 8.5 collisions (range 5-12), and MPC averaged 9 collisions per trial. ERPF's risk-ellipse inflation strategy successfully maintains collision-free operation by geometrically preempting worst-case trajectories.

Figure~\ref{ncp} depicts a complex highway scenario where the AV navigates among five HDVs with diverse behaviors: V2-V3 ahead potentially blocking progress, V4 behind limiting deceleration freedom, V5 contesting lane changes, and V6 approaching from the left front.

Figure~\ref{fig:highway_safety} validates ERPF's safety performance in this six-vehicle scenario. The spatial visualization shows repulsive forces (orange arrows) and the ego trajectory (red line) with evolution factor annotations. Key results include: (1) minimum clearance of 4.3 meters, well above safety thresholds; (2) zero collision events throughout the horizon; (3) smooth velocity profiles with stable longitudinal motion and bounded lateral adjustments, ensuring kinematic feasibility and comfort.

Figure~\ref{fig:runtime} evaluates computational performance for real-time applicability. The method demonstrates lightweight computation with 60.7 FLOPs per interaction (range 59-63), consistent 350 FLOPs per step, and 5,058 average FLOPs per timestep (peak 8,093). Total computational demand of 151,741 FLOPs per horizon is well within modern embedded system capabilities, confirming real-time feasibility for interaction-intensive scenarios.

\section{Conclusion}  
In this work, we proposed the ERPF-MPC framework and demonstrated its effectiveness in enhancing AV safety and adaptability in dynamic traffic environments. Through comparative analysis, we showed that ERPF provides a more responsive risk assessment, allowing the AV to better recognize and react to potential hazards while maintaining smooth and stable driving. Experimental results validated the superiority of ERPF over conventional RPF by highlighting its ability to increase field strength in critical moments without unnecessary interference in safe scenarios. The lane-changing and overtaking tests further confirmed the generalization capability of ERPF-MPC, enabling the AV to navigate dense traffic while minimizing collision risks. Overall, ERPF-MPC presents a significant improvement in motion planning by integrating adaptive risk evaluation, ensuring safer and more efficient autonomous driving. Future work will focus on expanding its applicability to more complex urban scenarios and further optimizing computational efficiency.

\section*{Funding}
{This work is sponsored by the Scientific Research Foundation of Chongqing University of Technology (0119240197). It was also supported by equipment funded through the “Intelligent Connected New Energy Vehicle Teaching System” project of Chongqing University of Technology, under the national initiative “Promote large-scale equipment renewals and trade-ins of consumer goods.” Furthermore, this paper was supported by Innovative Research Group of Chongqing Municipal Education Commission (CXQT19026), and Cooperative Project between Chinese Academy of Sciences and University in Chongqing (HZ2021011).}

\bibliographystyle{IEEEtran}
\bibliography{IEEEabrv,zq_lib}

\end{document}